\documentclass[10pt,twocolumn,letterpaper]{article}

 \usepackage{cvpr}              

\usepackage[dvipsnames]{xcolor}

\definecolor{cvprblue}{rgb}{0.21,0.49,0.74}
\usepackage[pagebackref,breaklinks,colorlinks,citecolor=cvprblue]{hyperref}
\usepackage{newfloat}
\usepackage{listings}
\usepackage{times}
\usepackage{epsfig}
\usepackage{graphicx}
\usepackage{amsmath}
\usepackage{amssymb}
\usepackage{graphicx}
\usepackage{amsmath}
\usepackage{amssymb}
\usepackage{booktabs}
\usepackage{bm}
\usepackage{multirow} 
\usepackage{algorithm}
\usepackage{algorithmic}
\usepackage{subcaption}
\usepackage{enumitem}
\usepackage{siunitx}

\usepackage{algorithm}
\usepackage{algorithmic}
\usepackage{algorithm}
\usepackage{algorithmic}
\usepackage{times}
\usepackage{epsfig}
\usepackage{graphicx}
\usepackage{amsmath}
\usepackage{amssymb}
\usepackage{amsmath,amssymb,amsfonts}
\usepackage{algorithm}
\usepackage{booktabs}
\usepackage{algorithmic}
\usepackage{bm}
\usepackage{multirow}
\usepackage{amssymb}
\usepackage{subcaption}
\usepackage{mathtools}
\usepackage{makecell}
\usepackage{enumitem}
\usepackage{caption}
\usepackage{subcaption}

\newcommand\blfootnote[1]{%
	\begingroup 
	\renewcommand\thefootnote{}\footnote{#1}%
	\addtocounter{footnote}{-1}%
	\endgroup 
}

\def\paperID{6416} 
\def\confName{CVPR}
\def\confYear{2024}

\title{CLIP-KD: An Empirical Study of CLIP Model Distillation}

\author{Chuanguang Yang$^{1,2}$ \qquad Zhulin An$^{1*}$ \qquad Libo Huang$^1$ \qquad Junyu Bi$^{1,2}$  \qquad Xinqiang Yu$^{1,2}$ \\ 
	Han Yang$^{1,2}$  \qquad Boyu Diao$^1$  \qquad Yongjun Xu$^{1*}$ \\
	$^1$Institute of Computing Technology, Chinese Academy of Sciences, Beijing, China  \\ $^2$University of Chinese Academy of Sciences, Beijing, China \\
	{\tt\small \{yangchuanguang, anzhulin, huanglibo, bijunyu, yuxinqiang21s\}@ict.ac.cn} \\
	{\tt\small \{yanghan22s, diaoboyu2012, xyj\}@ict.ac.cn} \\
}

\begin{document}
\maketitle

\begin{abstract}
		 Contrastive Language-Image Pre-training (CLIP) has become a promising language-supervised visual pre-training framework. This paper aims to distill small CLIP models supervised by a large teacher CLIP model. We propose several distillation strategies, including relation, feature, gradient and contrastive paradigms, to examine the effectiveness of CLIP-Knowledge Distillation (KD). We show that a simple feature mimicry with Mean Squared Error loss works surprisingly well. Moreover, interactive contrastive learning across teacher and student encoders is also effective in performance improvement. We explain that the success of CLIP-KD can be attributed to maximizing the feature similarity between teacher and student. The unified method is applied to distill several student models trained on CC3M+12M. CLIP-KD improves student CLIP models consistently over zero-shot ImageNet classification and cross-modal retrieval benchmarks. When using ViT-L/14 pretrained on Laion-400M as the teacher, CLIP-KD achieves 57.5\% and 55.4\% zero-shot top-1 ImageNet accuracy over ViT-B/16 and ResNet-50, surpassing the original CLIP without KD by 20.5\% and 20.1\% margins, respectively. Our code is released on \url{https://github.com/winycg/CLIP-KD}.
\end{abstract}

\section{Introduction}
\blfootnote{$^*$ Corresponding author}
Language-supervised image pre-training has attracted widespread attention for visual representation learning. As a representative work, CLIP (Contrastive Language-Image Pre-training)~\cite{radford2021learning} applies contrastive learning to (image, text) pairs. It guides the model to predict the correct (image, text) pair among the candidate image and text samples. Pre-trained CLIP models show excellent versatility in zero-shot multimodal and unimodal visual tasks.

Some recent works improve CLIP using an extra visual self-supervision task~\cite{mu2022slip,li2021supervision} or mask images~\cite{li2023scaling,yang2023attentive}. The pre-trained CLIP model is also introduced as a remarkable teacher to guide downstream visual pre-training~\cite{wei2022mvp,peng2022beit,kuo2022f}. However, few previous works explore improving the valuable small CLIP models in resource-constrained applications. This paper introduces CLIP-Knowledge Distillation (KD), which aims to enhance a small student CLIP model supervised by a pre-trained large teacher CLIP model. The state-of-the-art TinyCLIP~\cite{wu2023tinyclip} also investigates CLIP distillation. A critical core of TinyCLIP is weight inheritance, which transfers part weights from the well-trained teacher model to a smaller student model. However, this mechanism needs the teacher and student models to have the same architecture-style, \emph{e.g.}, ViT-B/32~\cite{dosovitskiy2020image} to ViT-61M/32 and ResNet-101~\cite{he2016deep} to ResNet-30M, limiting the scope of practical applications. This paper provides a comprehensive study on distilling small CLIP models from relation, feature, gradient, and contrastive paradigms. Our CLIP-KD does not rely on architectural-cue and can generalize to any teacher-student architecture pair.

Given the teacher and student CLIP models, we design distillation strategies from the view of mimicry and interaction. For mimicry learning, we guide the student to align the corresponding knowledge generated from the teacher, which is a basic framework in KD~\cite{hinton2015distilling,romero2014fitnets,beyer2022knowledge}. The core question is how to construct meaningful knowledge. Under CLIP, we build contrastive image-to-text relationships, (image, text) features and gradients for mimicry. For interactive learning, we combine the teacher and student for joint contrastive learning, letting the student learn from the teacher implicitly. For example, the student is regarded as an anchor to contrast the teacher embeddings. We also aggregate the student and teacher features for CLIP training.

We train CLIP models over Conceptual~\cite{sharma2018conceptual,changpinyo2021conceptual} datasets and evaluate pre-trained models over zero-shot ImageNet~\cite{deng2009imagenet} classification and cross-modal retrieval on MSCOCO~\cite{lin2014microsoft} and Flickr~\cite{young2014image}. All proposed distillation methods improve the student CLIP models with various margins. Surprisingly, a simple feature mimicry with Mean Squared Error loss could achieve the best performance. Moreover, interactive contrastive learning fulfills the second-best performance. We find that the distillation performance is in line with how much the feature similarity between teacher and student is maximized. This explains why various KD methods have different performance. The unified method is used to distill a series of student networks with different architectures and achieves consistent improvements. For example, when trained on CC3M+12M, CLIP-KD improves MobileViT-S from 32.6\% to 36.0\% on zero-shot ImageNet accuracy, reducing the gap with the teacher ViT-B/16's 37.0\%. When using ViT-L/14 pretrained on Laion-400M~\cite{schuhmann2021laion} as the teacher, CLIP-KD increases ViT-B/16 trained on CC3M+12M with a 20.5\% zero-shot ImageNet accuracy gain compared to the baseline.

The main contributions are summarized as follows:
\begin{itemize}
	\item We propose several distillation strategies, including relation, feature, gradient and contrastive paradigms, to examine the effectiveness of CLIP-KD.  A simple feature mimicry loss works surprisingly well. Interactive contrastive learning also achieves good performance.  
	\item We explain that a good CLIP distillation method could maximize the feature similarity between teacher and student models.  Intuitively, if
	the student's features perfectly align with the teacher's features, their performance gap could disappear. 
	\item We provide comprehensive guidelines for CLIP-KD. Compared to state-of-the-art TinyCLIP~\cite{wu2023tinyclip}, our CLIP-KD does not rely on architecture-cue and achieves better performance on both the same- and different-architecture styles of the teacher-student models.
\end{itemize}

\section{Related Works}
\textbf{Language-Supervised Learning.} Some previous multi-modal works explore visual representations supervised by language. A critical problem is how to create meaningful interaction between visual and linguistic. CLIP~\cite{radford2021learning} is a representative approach using contrastive learning over image-text pairs. ALIGN~\cite{jia2021scaling} utilizes larger-scale  contrastive pairs with noisy text supervision. Contrastive multi-modal learning~\cite{yuan2021multimodal,yu2022coca,fang2022eva,zhai2023sigmoid} has popularized exploring cross-modal correlation. Beyond the contrastive paradigm, generative approaches~\cite{desai2021virtex,wang2021simvlm,alayrac2022flamingo} have been examined for visual-linguistic learning. Our method focuses on CLIP distillation that improves the performance of the small CLIP models.

\textbf{CLIP Variants.} Some recent works attempt to improve CLIP with better performance and efficiency. SLIP~\cite{mu2022slip} combines CLIP and visual self-supervised learning as a multi-task framework. MaskCLIP~\cite{dong2023maskclip} introduces mask self-distillation to train an image EMA encoder for CLIP. DeCLIP~\cite{li2021supervision} performs data-efficient pre-training through multi-dimension supervision signals. Beyond auxiliary supervision, FLIP~\cite{li2023scaling} and A-CLIP~\cite{yang2023attentive} conduct image masking over the input to accelerate CLIP training and achieve a better trade-off between performance and efficiency. In contrast, our paper focuses on CLIP compression using KD instead of a new CLIP method.

\textbf{Multi-Modal Knowledge Distillation.} Knowledge Distillation (KD)~\cite{hinton2015distilling} has been applied to a broad range of tasks, such as visual recognition~\cite{li2021online,yang2021hierarchical,yang2022mixskd,li2023curriculum,huang2023etag}, language model compression~\cite{jiao2019tinybert}, and multi-modal representation learning~\cite{fang2021compressing,li2024promptkd}. DistillVLM~\cite{fang2021compressing} aligns hidden attention distributions and feature maps between teacher and student. This simple yet effective idea has been applied to many multi-modal KD works~\cite{wang2022multimodal,li2023distilling,liang2024module}. Recently, TinyCLIP~\cite{wu2023tinyclip} also aims for CLIP distillation and achieves satisfactory performance via affinity mimicking and weight inheritance. However,  the weight inheritance mechanism requires the same architecture-style between teacher and student models. By contrast, CLIP-KD could adapt any architecture pair without considering architectural correlation.

\begin{figure*}[t]
	\centering 
	
	\begin{subfigure}[t]{0.495\textwidth}
		\centering
		\includegraphics[width=\textwidth]{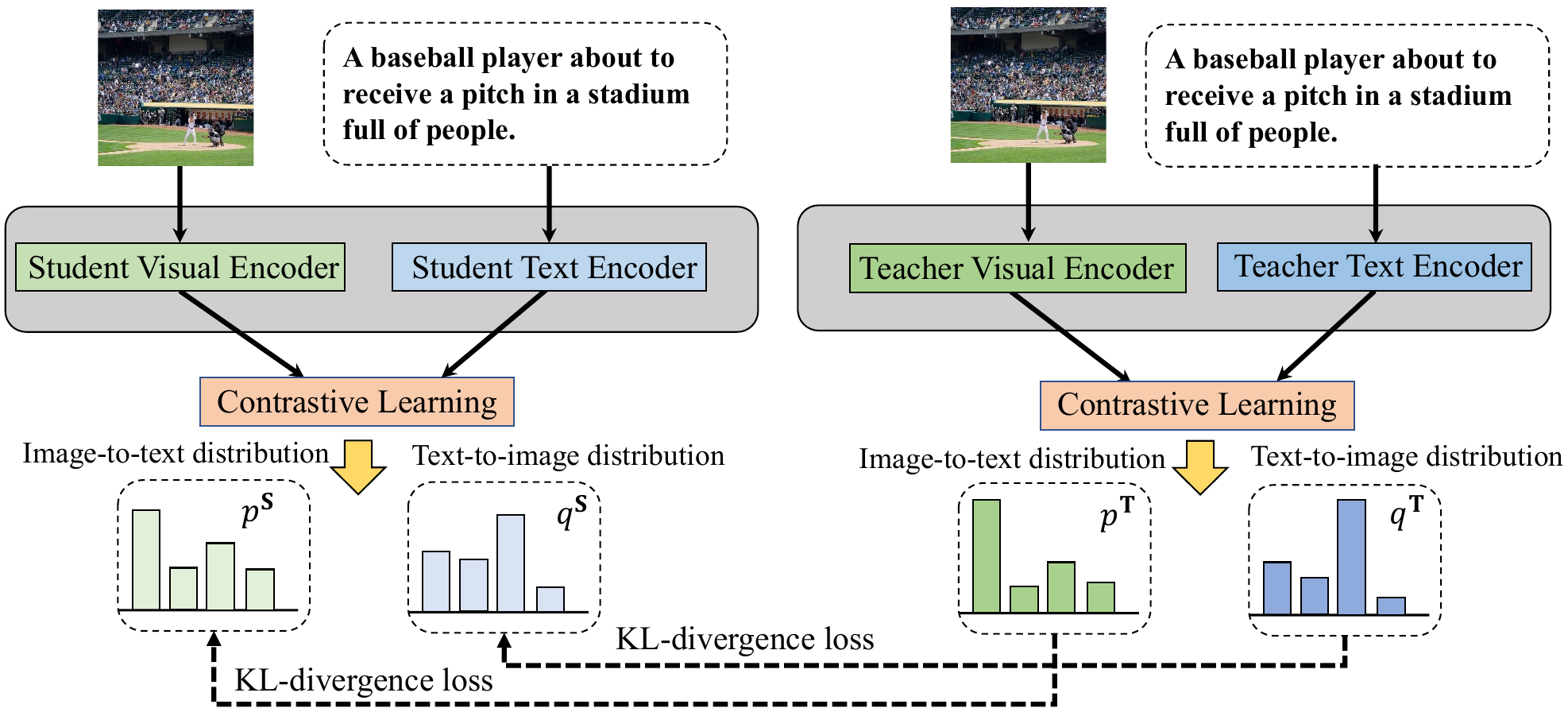}
		\caption{Contrastive Relational Distillation (CRD).}
		\label{CRD}
	\end{subfigure}
	\begin{subfigure}[t]{0.495\textwidth}
		\centering
		\includegraphics[width=\textwidth]{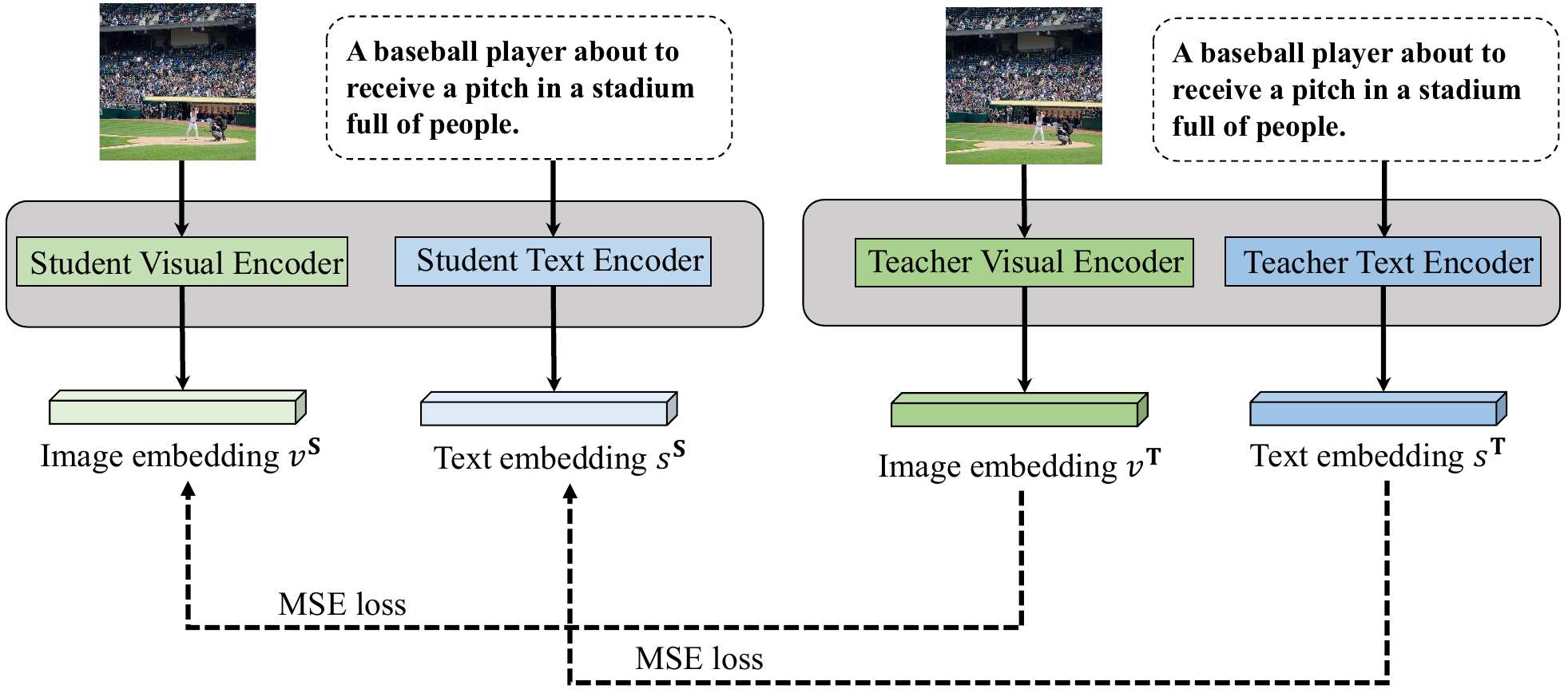}
		\caption{Feature Distillation (FD).}
		\label{FD}
	\end{subfigure}
	\begin{subfigure}[t]{0.495\textwidth}
		\centering
		\includegraphics[width=\textwidth]{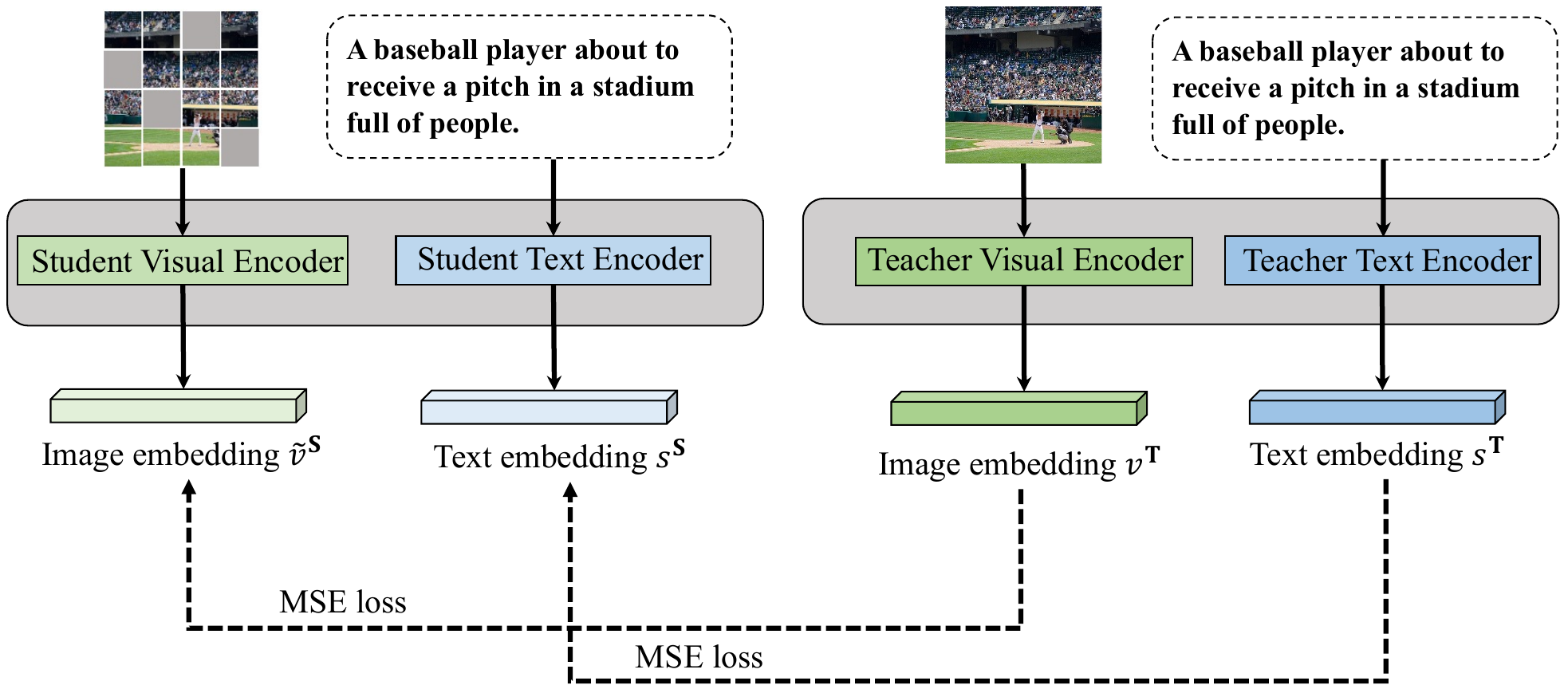}
		\caption{Masked Feature Distillation (MFD).}
		\label{MFD}
	\end{subfigure}
	\begin{subfigure}[t]{0.495\textwidth}
		\centering
		\includegraphics[width=\textwidth]{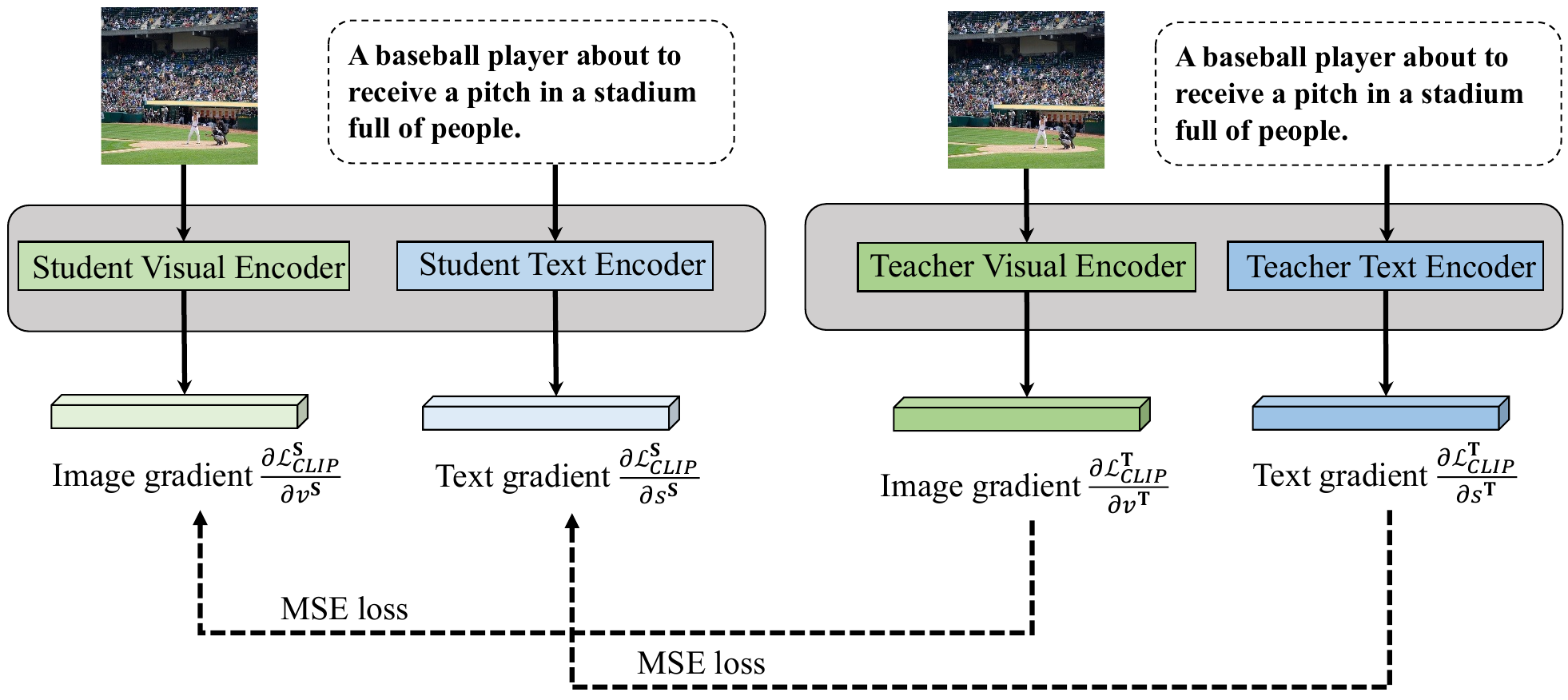}
		\caption{Gradient Distillation (GD).}
		\label{GD}
	\end{subfigure}
	\begin{subfigure}[t]{0.498\textwidth}
		\centering
		\includegraphics[width=\textwidth]{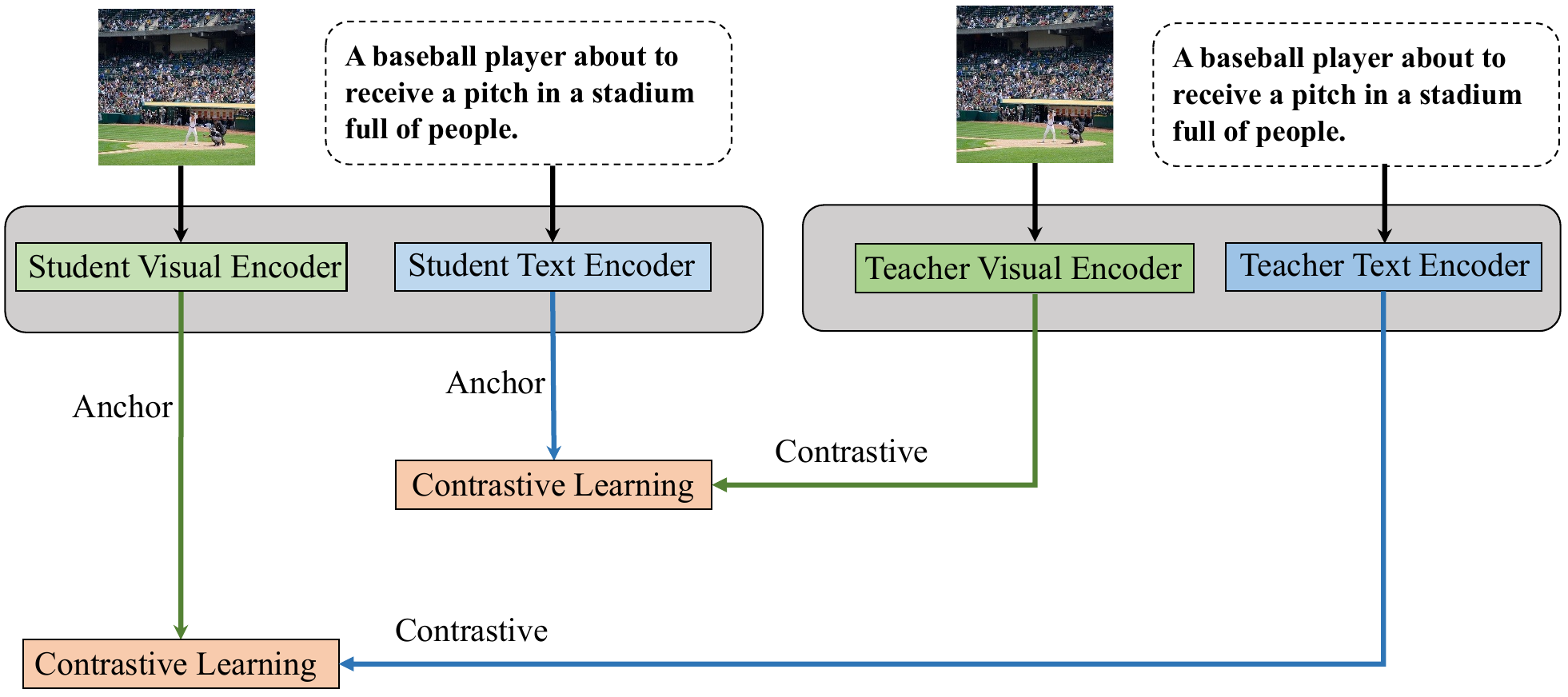}
		\caption{Interactive Contrastive Learning (ICL).}
		\label{ICL}
	\end{subfigure}
	\begin{subfigure}[t]{0.493\textwidth}
		\centering
		\includegraphics[width=\textwidth]{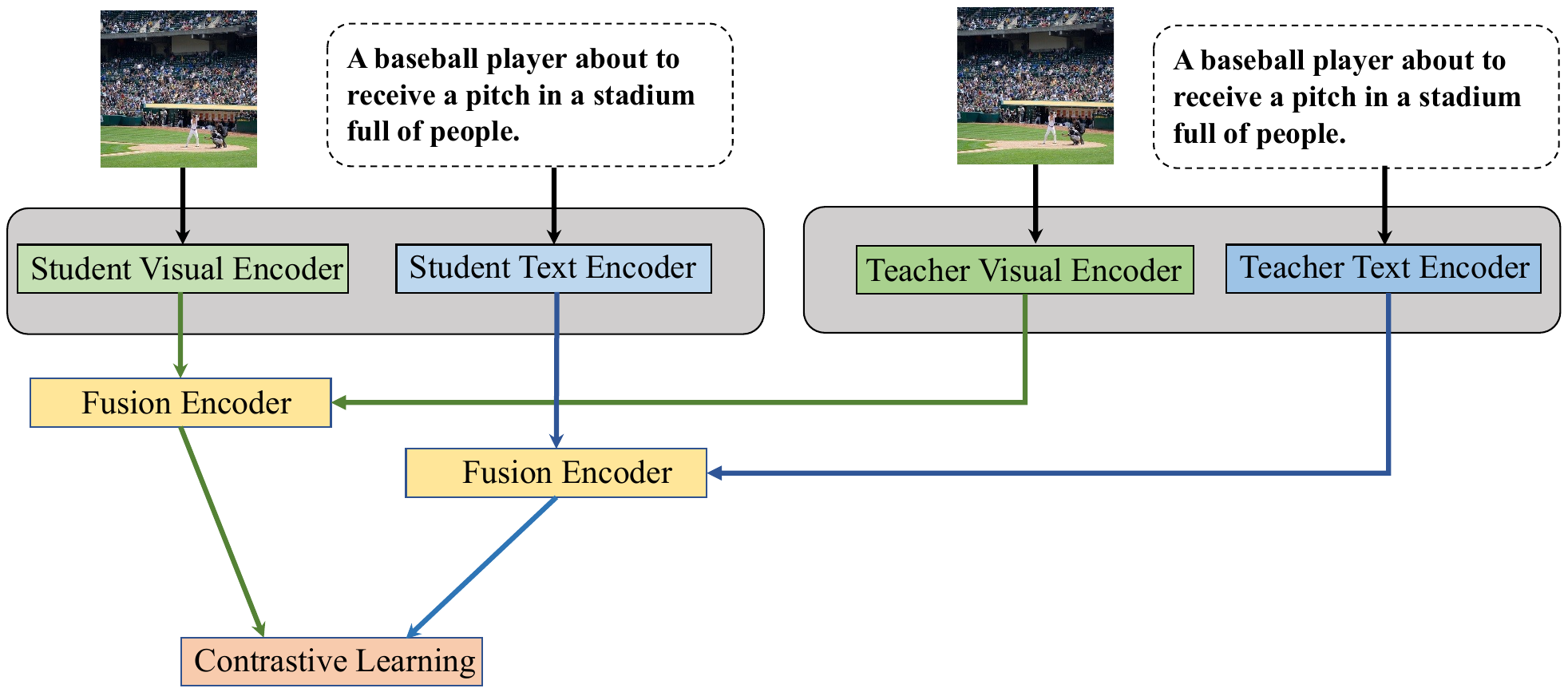}
		\caption{Augmented Feature Distillation (AFD).}
		\label{AFD}
	\end{subfigure}
	
	\caption{Illustration of various CLIP knowledge distillation approaches proposed in this paper.} 
	\label{clip_kd} 
	\vspace{-0.4cm}
\end{figure*}

\section{Methodology}
\subsection{A Brief Review of CLIP}
\textbf{CLIP (Contrastive Language-Image Pre-Training).} Given a set of (image, text) pairs denoted as $\mathcal{D}=\{(I_{k},T_{k})\}_{k=1}^{|\mathcal{D}|}$, CLIP performs an image-text alignment task to push the paired image-text close and unpaired ones apart in the feature embedding space. The CLIP framework includes a visual encoder $f_{i}$ and a text encoder $f_{t}$ to transform the image $I_{k}$ and the text $T_{k}$ into feature embeddings $v_{k}$ and $s_{k}$ respectively, \emph{i.e.} $v_{k}=f_{i}(I_{k})$, $s_{k}=f_{t}(T_{k})$. Here, all embeddings are post-processed by $l_2$ normalization.  CLIP adopts InfoNCE-based~\cite{oord2018representation} contrastive loss to maximize the similarity between $v_{k}$ and $s_{k}$ against other negative samples. Given the image embedding $v_{k}$ as the anchor, the image-to-text contrastive loss is formulated as:
\begin{equation}
	\mathcal{L}_{I\to T}=-\log\frac{\exp(v_{k}\cdot s_{k}/\tau)}{\sum_{b=1}^{|\mathcal{B}|}\exp(v_{k}\cdot s_{b}/\tau)}.
\end{equation}
CLIP conducts a symmetric image-text alignment contrastive loss. Given the text embedding $s_{k}$ as the anchor, the text-to-image contrastive loss is formulated as:
\begin{equation}
	\mathcal{L}_{T\to I}=-\log\frac{\exp(s_{k}\cdot v_{k}/\tau)}{\sum_{b=1}^{|\mathcal{B}|}\exp(s_{k}\cdot v_{b}/\tau)}.
\end{equation}
Here, $\cdot$ denotes the dot product to measure the similarity, $\tau$ is a learnable temperature to scale the distribution. In practice, the negative samples are retrieved from the mini-batch $\mathcal{B}$. The total loss of CLIP is formulated as:
\begin{equation}
	\mathcal{L}_{CLIP}=\frac{1}{2}(\mathcal{L}_{I\to T}+\mathcal{L}_{T\to I}).
	\label{clip}
\end{equation}

\subsection{CLIP Knowledge Distillation}
In this section, we propose several CLIP distillation methods and illustrate the overview of details in Fig.~\ref{clip_kd}.
\label{CLIP_KD}
\subsubsection{Contrastive Relational Distillation}
The core idea of CLIP is to maximize the similarity between the paired image-text embeddings over the contrastive similarity distribution. Therefore, the straightforward knowledge type is output-oriented contrastive distribution for Contrastive Relational Distillation (CRD). This idea is also used by some previous works for image classification~\cite{fang2021seed,yang2022mutual,yang2023online}, object detection~\cite{yao2021g} and semantic segmentation~\cite{yang2022cross}. The contrastive distribution captures the structured relationships among feature embeddings. A good teacher often constructs a well-structured feature space. CRD makes the student mimic better structured semantic relations from the teacher, further improving the quality of feature representations.

Given a mini-batch $\mathcal{B}=\{(I_{k},T_{k})\}_{k=1}^{|\mathcal{B}|}$, the generated (visual, text) embeddings from teacher and student are $\{(v_{k}^{\mathbf{T}},s_{k}^{\mathbf{T}})\}_{k=1}^{|\mathcal{B}|}$ and  $\{(v_{k}^{\mathbf{S}},s_{k}^{\mathbf{S}})\}_{k=1}^{|\mathcal{B}|}$, respectively. Given the $k$-th image embedding $v_{k}$ as an anchor, the teacher and student image-to-text contrastive distributions $p_{k}^{\mathbf{T}}\in \mathbb{R}^{|\mathcal{B}|}$ and $p_{k}^{\mathbf{S}}\in \mathbb{R}^{|\mathcal{B}|}$ are formulated as:
\begin{equation}
	p_{k}^{\mathbf{T}}[j]=\frac{\exp(v_{k}^{\mathbf{T}}\cdot s_{j}^{\mathbf{T}}/\tau^{\mathbf{T}})}{\sum_{b=1}^{|\mathcal{B}|}\exp(v_{k}^{\mathbf{T}}\cdot s_{b}^{\mathbf{T}}/\tau^{\mathbf{T}})},
	\label{p}
\end{equation}
\begin{equation}
	p_{k}^{\mathbf{S}}[j]=\frac{\exp(v_{k}^{\mathbf{S}}\cdot s_{j}^{\mathbf{S}}/\tau^{\mathbf{S}})}{\sum_{b=1}^{|\mathcal{B}|}\exp(v_{k}^{\mathbf{S}}\cdot s_{b}^{\mathbf{S}}/\tau^{\mathbf{S}})}.
\end{equation}
Here, $j\in [1,2,\cdots,|\mathcal{B}|]$ denotes the index of the contrastive distribution. Symmetrically, given the text embedding $s_{k}$ as an anchor, the teacher and student text-to-image contrastive distributions $q_{k}^{\mathbf{T}}\in \mathbb{R}^{|\mathcal{B}|}$ and $q_{k}^{\mathbf{S}}\in \mathbb{R}^{|\mathcal{B}|}$ are formulated as:
\begin{equation}
	q_{k}^{\mathbf{T}}[j]=\frac{\exp(s_{k}^{\mathbf{T}}\cdot v_{j}^{\mathbf{T}}/\tau^{\mathbf{T}})}{\sum_{b=1}^{|\mathcal{B}|}\exp(s_{k}^{\mathbf{T}}\cdot v_{b}^{\mathbf{T}}/\tau^{\mathbf{T}})},
\end{equation}
\begin{equation}
	q_{k}^{\mathbf{S}}[j]=\frac{\exp(s_{k}^{\mathbf{S}}\cdot v_{j}^{\mathbf{S}}/\tau^{\mathbf{S}})}{\sum_{b=1}^{|\mathcal{B}|}\exp(s_{k}^{\mathbf{S}}\cdot v_{b}^{\mathbf{S}}/\tau^{\mathbf{S}})}.
\end{equation}

We align the contrastive distributions between teacher and student via KL-divergence loss. For image-to-text and text-to-image, the distillation losses are formulated as Eq.(\ref{CRD_I}) and Eq.(\ref{CRD_T}):
\begin{equation}
	\mathcal{L}_{CRD\_I\to T}=\frac{1}{|\mathcal{B}|}\sum_{k=1}^{|\mathcal{B}|}\sum_{j=1}^{|\mathcal{B}|}p_{k}^{\mathbf{T}}[j]\log{\frac{p_{k}^{\mathbf{T}}[j]}{p_{k}^{\mathbf{S}}[j]}},
	\label{CRD_I}
\end{equation}
\begin{equation}
	\mathcal{L}_{CRD\_T\to I}=\frac{1}{|\mathcal{B}|}\sum_{k=1}^{|\mathcal{B}|}\sum_{j=1}^{|\mathcal{B}|}q_{k}^{\mathbf{T}}[j]\log{\frac{q_{k}^{\mathbf{T}}[j]}{q_{k}^{\mathbf{S}}[j]}}.
	\label{CRD_T}
\end{equation}
The total CRD loss for CLIP distillation is summarized as:
\begin{equation}
	\mathcal{L}_{CRD}=\mathcal{L}_{CRD\_I\to T}+\mathcal{L}_{CRD\_T\to I}.
\end{equation}

\subsubsection{Feature Distillation}
A simple yet effective way is to align feature embeddings between teacher and student to reduce the knowledge gap directly.  Intuitively, if the student's features perfectly align with the teacher's features, their performance gap could disappear. We guide the student to mimic the teacher's visual and text embeddings via Mean Squared Error (MSE) loss:
\begin{equation}
	\mathcal{L}_{FD}=\frac{1}{|\mathcal{B}|}\sum_{k=1}^{|\mathcal{B}|}(\left \| v_{k}^{\mathbf{T}}-v_{k}^{\mathbf{S}} \right \|  ^{2}_{2}+\left \| s_{k}^{\mathbf{T}}-s_{k}^{\mathbf{S}} \right \|  ^{2}_{2}).
\end{equation}
Here, when the embedding sizes between teacher and student are different, we apply a linear projection head to student embeddings to match the dimension.
\subsubsection{Masked Feature Distillation}
The core idea of masked image modeling~\cite{bao2021beit,xie2022simmim,he2022masked} is to recover the masked regions using contextual information modeling by a vision transformer. In the scenario of distillation, the teacher is a good supervisor that could provide valuable information to help the student recover the visual semantics given the masked image as input. Like FD, we utilize MSE loss to align the student's and teacher's visual and text embeddings. The difference is that Masked Feature Distillation (MFD) uses masked images as the input to a student. The patch masking algorithm is followed from MAE~\cite{he2022masked}. The total loss of MFD is formulated as:
\begin{equation}
	\mathcal{L}_{MFD}=\frac{1}{|\mathcal{B}|}\sum_{k=1}^{|\mathcal{B}|}(\left \| v_{k}^{\mathbf{T}}-\tilde{v}_{k}^{\mathbf{S}} \right \|  ^{2}_{2}+\left \| s_{k}^{\mathbf{T}}-s_{k}^{\mathbf{S}} \right \|  ^{2}_{2}),
\end{equation}
where $\tilde{v}_{k}^{\mathbf{S}}$ is the visual embedding based on the masked input image.

\subsubsection{Gradient Distillation}
The gradient information often shows
how the model responds to changes according to inputs. We propose to force the gradient consistency between teacher and student using the derivative w.r.t the visual and text embeddings. By this means, the student could better understand how the output should change according to the input. This helps the student behave more similarly to the teacher.

We align the gradient information w.r.t each visual and text embedding between teacher and student via MSE loss:
\begin{align}
	\label{grad}
	\mathcal{L}_{GD}=\frac{1}{|\mathcal{B}|}\sum_{k=1}^{|\mathcal{B}|}&(\left \| \frac{\partial \mathcal{L}_{CLIP}^{\mathbf{T}}}{\partial v_k^{\mathbf{T}}}-\frac{\partial \mathcal{L}_{CLIP}^{\mathbf{S}}}{\partial v_k^{\mathbf{S}}} \right \|  ^{2}_{2} \notag\\
	&+\left \| \frac{\partial \mathcal{L}_{CLIP}^{\mathbf{T}}}{\partial s_k^{\mathbf{T}}}-\frac{\partial \mathcal{L}_{CLIP}^{\mathbf{S}}}{\partial s_k^{\mathbf{S}}} \right \|  ^{2}_{2}).	
\end{align}
\emph{Derivations in Eq.(\ref{grad}) are shown in Appendix Section 1.}

\subsubsection{Interactive Contrastive Learning}
To facilitate the interaction between teacher and student, we propose \emph{Interactive Contrastive Learning} (ICL) across the student and teacher feature encoders. It regards the student as an anchor to contrast the teacher's embeddings. Given the student image embedding $v_k^{\mathbf{S}}$, the contrastive text embeddings denoted as $\{s_{b}^{\mathbf{T}}\}_{b=1}^{|\mathcal{B}|}$ are from the teacher text encoder instead of the student text encoder. The image-to-text ICL loss is formulated as:
\begin{equation}
	\mathcal{L}_{ICL\_I\to T}=-\log\frac{\exp(v_{k}^{\mathbf{S}}\cdot s_{k}^{\mathbf{T}}/\tau)}{\sum_{b=1}^{|\mathcal{B}|}\exp(v_{k}^{\mathbf{S}}\cdot s_{b}^{\mathbf{T}}/\tau)}.
\end{equation}
Symmetrically, given the student text embedding $s_k^{\mathbf{S}}$, the contrastive image embeddings denoted as $\{v_{b}^{\mathbf{T}}\}_{b=1}^{|\mathcal{B}|}$ are from the teacher visual encoder. The text-to-image ICL loss is formulated as:
\begin{equation}
	\mathcal{L}_{ICL\_T\to I}=-\log\frac{\exp(s_{k}^{\mathbf{S}}\cdot v_{k}^{\mathbf{T}}/\tau)}{\sum_{b=1}^{|\mathcal{B}|}\exp(s_{k}^{\mathbf{S}}\cdot v_{b}^{\mathbf{T}}/\tau)}.
\end{equation}
The total loss of ICL is summarized as:
\begin{equation}
	\mathcal{L}_{ICL}=\frac{1}{2}(\mathcal{L}_{ICL\_I\to T}+\mathcal{L}_{ICL\_T\to I}).
\end{equation}
We demonstrate that minimizing $\mathcal{L}_{ICL}$ is connected to maximizing the lower bound of mutual information between teacher and student networks. The mutual information measures the uncertainty reduction in contrastive feature embeddings from the teacher network when the anchor embedding from the student network is known. By maximizing the lower bound of mutual information, the student network learns more common knowledge from the teacher network, leading to better feature representations. \emph{The theoretical proof is shown in Appendix Section 2.}

\begin{table*}[tbp]
	\centering
	\caption{\textbf{Comparison of CLIP distillation losses trained from CC3M+12M on zero-shot ImageNet-related classification and cross-modal retrieval on CC3M Val, MSCOCO and Flickr.} The numbers in \textbf{bold} denote the best results for individual methods (the third block) and unified methods (the fourth block), respectively. The 'T' and 'S' tags represent the teacher and student roles, respectively.}
	\vspace{-0.2cm}
	\resizebox{0.85\linewidth}{!}{
		\begin{tabular}{l|cccc|cc|cc|cc}  
			\toprule
			\multirow{2}{*}{Method}& IN&INV2&IN-R&IN-S&\multicolumn{2}{c|}{CC3M Val}&\multicolumn{2}{c|}{MSCOCO}&\multicolumn{2}{c}{Flickr}\\ 
			&Acc&Acc&Acc&Acc&I2T&T2I&I2T&T2I&I2T&T2I\\
			\midrule
			T: ViT-B/16 
			&37.0 &32.1&48.4&26.0&40.2&39.5&25.0&24.7&54.6&56.6\\
			\midrule
			S: ViT-T/16 	&30.6&25.6&35.7&17.4&33.3&33.3&20.7&20.3&46.4&47.7\\
			+CRD& 31.9&27.6&38.8&19.6&35.3&34.9&21.4&20.7&48.8&49.9\\
			+FD&\textbf{34.2}&\textbf{29.5}&\textbf{42.7}&\textbf{21.4}&37.1&\textbf{36.9}&22.5&\textbf{22.2}&\textbf{51.1}&\textbf{51.3}\\
			+MFD& 34.1&\textbf{29.5}&42.3&21.2&\textbf{37.4}&\textbf{36.9}&\textbf{22.9}&22.1&50.9&51.1\\
			+GD& 31.5&27.0&37.9&19.1&34.5&34.0&21.3&20.9&47.5&48.3\\
			+ICL& 33.1&28.2&40.6&20.8&36.1&35.8&21.8&21.7&50.5&50.4\\
			+AFD& 31.4&26.9&37.8&18.6&34.6&34.7&20.9&20.5&47.3&48.7\\
			\midrule
			+FD+ICL &34.6&30.0&43.2&\textbf{22.0}&37.9&37.6&23.0&22.5&51.7&51.9\\
			
			+FD+ICL+CRD&\textbf{34.9}&\textbf{30.1}&43.5&21.9&\textbf{38.2}&\textbf{37.9}&23.1&\textbf{22.6}&52.3&\textbf{52.4}\\
			+FD+ICL+CRD+GD &34.8&29.9&42.8&\textbf{22.0}&38.1&37.7&\textbf{23.3}&22.5&\textbf{52.4}&52.3\\
			+FD+ICL+CRD+AFD &34.8&\textbf{30.1}&\textbf{43.6}&21.6&\textbf{38.2}&37.7&23.0&22.5&52.2&\textbf{52.4}\\
			\bottomrule
	\end{tabular}}
	\label{comparison_clip} 
	\vspace{-0.4cm}
\end{table*}

\subsubsection{Augmented Feature Distillation}
To help the interaction between teacher and student, we propose to augment the student embeddings using the teacher embeddings by a fusion encoder. We hope the teacher can guide the student to optimize a meaningful visual-text embedding space. We introduce a visual fusion encoder $\phi_i$ and a text fusion encoder $\phi_t$ to aggregate the student and teacher embeddings:
\begin{equation}
	v_{k}^{\mathbf{A}}=\phi_i(v_{k}^{\mathbf{S}}||v_{k}^{\mathbf{T}}), 	s_{k}^{\mathbf{A}}=\phi_t(s_{k}^{\mathbf{S}}||s_{k}^{\mathbf{T}}).
\end{equation}
Here, $||$ is the concatenation operator, and the fusion encoder is a simple linear projection layer. The augmented feature embeddings $(v_{k}^{\mathbf{A}},s_{k}^{\mathbf{A}})$ are applied to the general CLIP contrastive loss as Eq.(\ref{clip}).

\subsubsection{Overall Loss of CLIP Distillation.}
We summarize the original CLIP task and distillation losses to jointly train a student model:
\begin{equation}
	\mathcal{L}_{CLIP\_KD}=\mathcal{L}_{CLIP}+\lambda \mathcal{L}_{KD}.
\end{equation}
Here, $\mathcal{L}_{KD}\in \{\mathcal{L}_{CRD},\mathcal{L}_{FD},\mathcal{L}_{MFD},\mathcal{L}_{GD},\mathcal{L}_{ICL},\mathcal{L}_{AFD}\}$ represents a distillation loss. $\lambda$ is a distillation loss weight to scale the magnitude. Multiple distillation losses can be selectively utilized together.

\section{Experiments}
\subsection{Experimental Setup}
\textbf{Dataset.} We use Conceptual Captions 3M (CC3M)~\cite{sharma2018conceptual} and Conceptual 12M (CC12M)~\cite{changpinyo2021conceptual} for vision-and-language pre-training. We follow the consistent evaluation protocol with CLIP-related works~\cite{wei2022icar,li2023scaling}. The CC3M validation set, including 13K image-text pairs, is used for cross-modal retrieval evaluation. For zero-shot classification, we utilize the ImageNet (IN)~\cite{deng2009imagenet} validation set and its several variants, such as ImageNet-V2 (IN-V2)~\cite{recht2019imagenet}, ImageNet-Rendition (IN-R)~\cite{hendrycks2021many} and ImageNet-Sketch (IN-S)~\cite{wang2019learning} for evaluation. For zero-shot cross-modal image/text retrieval, we adopt MSCOCO~\cite{lin2014microsoft} and Flickr~\cite{young2014image} for evaluation.

\textbf{Evaluation metrics.} Following the standard setting, we employ Recall@$K$ (R@$K$) to measure the retrieval performance in $K$ nearest neighbours. By default, we use top-1 accuracy (Acc) for image classification and R@1 for Image-to-Text (I2T) and Text-to-Image (T2I) retrieval.

\begin{table}[tbp]
	\centering
	\caption{Configuration of paired visual and text encoders.}
		\vspace{-0.2cm}
	\resizebox{1.\linewidth}{!}{
		\begin{tabular}{l|c|c|c|c|c|c}  
			\toprule
			\multicolumn{3}{c|}{Visual encoder} & \multicolumn{4}{c}{Text encoder: Transformer~\cite{vaswani2017attention}} \\
			\hline
			Model & Type & Params & Layer & Width & Head & Params \\
			\midrule
			ViT-L/14~\cite{dosovitskiy2020image} & \multirow{5}{*}{ViT} & 304.0M & 12 & 768 & 12 & 85.1M \\
			ViT-B/16~\cite{dosovitskiy2020image} &  & 86.2M & 12 & 512 & 8 & 37.8M \\
			\cline{4-7} 
			ViT-T/16~\cite{dosovitskiy2020image} &  & 5.6M & \multirow{3}{*}{12}  & \multirow{3}{*}{384}  & \multirow{3}{*}{6}  & \multirow{3}{*}{21.3M} \\
			MobileViT-S~\cite{mehta2021mobilevit}&  & 5.3M &  &  &  &  \\
			Swin-T~\cite{liu2021swin}&  & 27.9M &  &  &  &  \\
		\midrule
			ResNet-101~\cite{he2016deep},&  \multirow{5}{*}{CNN}  &56.3M & \multirow{2}{*}{12} & \multirow{2}{*}{512} &  \multirow{2}{*}{8} &  \multirow{2}{*}{37.8M} \\
			ResNet-50~\cite{he2016deep},&  & 38.3M&  &  &  &  \\
			\cline{4-7} 
			ResNet-18~\cite{he2016deep},&  & 11.4M&  \multirow{3}{*}{12}  & \multirow{3}{*}{384}  & \multirow{3}{*}{6}  & \multirow{3}{*}{21.3M}  \\
			MobileNetV3~\cite{howard2019searching}&  & 2.0M&  &  &  &  \\
			EfficientNet-B0~\cite{tan2019efficientnet} & & 4.7M &  &  &  &  \\
			\bottomrule
			
	\end{tabular}}
	
	\label{config} 
	\vspace{-0.5cm}
\end{table}

\begin{table*}[tbp]
	\centering
	\caption{Distillation performance  trained from CC3M+12M  for cross-modal retrieval on CC3M, MSCOCO and Flickr validation set. }
	\vspace{-0.2cm}
	\resizebox{1.\linewidth}{!}{
		\begin{tabular}{l|cc|cc|cc||l|cc|cc|cc}  
			\toprule
			\multirow{2}{*}{Method}& \multicolumn{2}{c|}{CC3M}&\multicolumn{2}{c|}{MSCOCO}&\multicolumn{2}{c||}{Flickr}&\multirow{2}{*}{Method} &\multicolumn{2}{c|}{CC3M}&\multicolumn{2}{c|}{MSCOCO}&\multicolumn{2}{c}{Flickr}\\ 
			&I2T&T2I&I2T&T2I&I2T&T2I&&I2T&T2I&I2T&T2I&I2T&T2I\\
			\midrule
			T: ViT-B/16 
			&40.2&39.5&25.0 &24.7&54.6&56.6& T: ResNet-101 
			&41.4&40.5&25.2&25.7&57.0&55.5\\
			\midrule
			S: MobileViT-S 	&36.0&35.6&22.3&22.9&50.1&53.0&S: MobileViT-S 	&36.0&35.6&22.3&22.9&50.1&53.0\\
			+CLIP-KD& \textbf{39.4}&\textbf{38.6}&\textbf{26.1}&\textbf{24.9}&\textbf{55.0}&\textbf{56.2}&
			+CLIP-KD& \textbf{39.9}&\textbf{38.6}&\textbf{26.0}&\textbf{25.3}&\textbf{57.6}&\textbf{56.1}\\
			\midrule
			S: Swin-T 	&39.8&39.2&24.7&25.3&53.4&54.4
			&S: Swin-T 	&39.8&39.2&24.7&25.3&53.4&54.4 \\
			+CLIP-KD& \textbf{43.7}&\textbf{42.5}&\textbf{28.5}&\textbf{28.6}&\textbf{62.2}&\textbf{60.9}&
			+CLIP-KD& \textbf{44.2}&\textbf{43.0}&\textbf{27.8}&\textbf{28.9}&\textbf{60.8}& \textbf{61.5 }\\
			\midrule
			S: MobileNetV3 	&28.1&27.5&15.3&15.0&36.9&38.0  &S: MobileNetV3 	&28.1&27.5&
			15.3&15.0&36.9&38.0 \\
			+CLIP-KD& \textbf{30.1}&\textbf{28.6}&\textbf{17.9}&\textbf{16.0}&\textbf{42.4}&\textbf{42.3}&+CLIP-KD& \textbf{30.2}&\textbf{29.4}&\textbf{17.2}&\textbf{16.6}&\textbf{40.2} &\textbf{42.2}\\
			\midrule
			S: EfficientNet-B0 	&35.4&34.9&
			21.7&21.1&48.3&50.1&S: EfficientNet-B0 	&35.4&34.9&
			21.7&21.1&48.3&50.1\\
			+CLIP-KD&\textbf{39.0}&\textbf{38.0}&\textbf{26.0}&\textbf{23.9}&\textbf{55.5}&\textbf{54.2}&+CLIP-KD&\textbf{37.4}&\textbf{36.8}&\textbf{24.7}&\textbf{24.6} &\textbf{55.8}& \textbf{56.2}\\
			\midrule
			S: ResNet-18 	&31.1&30.4&19.2&18.6 &41.0&43.3&S: ResNet-18 	&31.1&30.4&19.2&18.6 &41.0&43.3\\
			+CLIP-KD&\textbf{34.2}&\textbf{33.0}&\textbf{21.3}&\textbf{19.8}&\textbf{47.8}&\textbf{47.1}&+CLIP-KD&\textbf{34.7}&\textbf{33.7}&\textbf{21.0}&\textbf{20.9}&\textbf{48.8} &\textbf{48.4}\\
			\bottomrule
	\end{tabular}}
	
	\label{vit_b_kd_cc3m_val} 
	\vspace{-0.5cm}
\end{table*}

\textbf{Configuration of visual and text encoders.} As shown in Table~\ref{config}, We show the configuration of visual and text encoders, followed by open\_clip\footnote{https://github.com/mlfoundations/open\_clip} codebase.

\textbf{Training details.} We adopt an AdamW optimizer~\cite{loshchilov2017decoupled} with an initial learning rate of 0.001 and a weight decay of 0.1. A cosine learning rate schedule is applied with a linear warm-up for 10K iterations in 32 epochs. Experiments are run over 8 NVIDIA A800 GPUs. The batch size is 1024, where each GPU holds 128 samples. For the weight of each distillation loss, we set $\lambda_{CRD}=1$, $\lambda_{FD}=\lambda_{MFD}=2000$, $\lambda_{GD}=10^{8}$ and $\lambda_{ICL}=1$. The learnable temperature $\tau$ is initialized from 0.07. Other training settings are followed from the original CLIP~\cite{radford2021learning}. \emph{The detailed hyper-parameter experiments  are shown in Appendix Section 3.}

\subsection{Ablation Study of Distillation Losses}
In this section, we examine the effectiveness of various CLIP distillation approaches. As shown in Table~\ref{comparison_clip}, we conduct a comprehensive comparison on zero-shot ImageNet-related classification and cross-modal retrieval. Any individual distillation loss could boost the student performance over the baseline. Feature Distillation (FD) with a simple MSE mimicry loss achieves the best distillation performance among them. It improves the student by 3.6\% top-1 accuracy on ImageNet, 3.7\%, 1.9\% and 4.2\% R@1 values on CC3M, MSCOCO and Flickr, respectively. We further evaluate MFD by applying image patch masking into FD. MFD shows similar performance with FD, therefore we do not introduce this technique for CLIP-KD. Beyond MFD, ICL and CRD become the second- and third-best approaches for overall zero-shot performance. GD and AFD lead to relatively moderate performance gains compared to the baseline. 

We further combine loss terms to investigate the unified distillation approach. The combination of FD+ICL outperforms the single FD or ICL, indicating that FD and ICL are complementary.  We further apply CRD to FD+ICL, and the performance is improved continually. Moreover, we find adding GD or AFD to FD+ICL+CRD may not lead to performance gains. In summary, the combination FD+CRD+ICL performs best in 6 out of 10 cases. By default, we utilize this unified method for distilling various CLIP models in this paper.

\begin{table*}[tbp]
	\centering
	\caption{Distillation performance of zero-shot ImageNet and its variants on top-1 classification accuracy (\%)  trained on CC3M+12M.}
	\vspace{-0.2cm}
	\resizebox{0.85\linewidth}{!}{
		\begin{tabular}{l|cccc||l|cccc}  
			\toprule
			Method&IN-1K&INV2& IN-R&IN-S&Method&IN-1K&INV2& IN-R&IN-S\\ 
			\midrule
			T: ViT-B/16 
			&37.0&32.1&48.4&26.0 &T: ResNet-101 
			&36.8 &31.9&49.2&26.7\\
			\midrule
			S: MobileViT-S  &32.6  & 27.6&39.5&20.1&S: MobileViT-S &32.6 & 27.6&39.5&20.1 \\
			+CLIP-KD& \textbf{36.0} &\textbf{31.1}&\textbf{44.5}&\textbf{23.5}&+CLIP-KD& \textbf{35.0} & \textbf{30.1} &\textbf{43.7}&\textbf{22.7}\\
			\midrule
			S: Swin-T& 36.4 & 31.1&45.9&24.4&S: Swin-T& 36.4 & 31.1&45.9&24.4\\
			+CLIP-KD&\textbf{40.2}&\textbf{34.9}&\textbf{51.4}&\textbf{28.2}&+CLIP-KD&\textbf{39.5}&\textbf{34.2} &\textbf{51.9} &\textbf{28.1} \\ 
			\midrule
			S: MobileNetV3& 25.1 & 20.7&29.2&13.4&S:MobileNetV3& 25.1& 20.7&29.2&13.4\\
			+CLIP-KD&\textbf{27.0}&\textbf{23.0}&\textbf{30.6}&\textbf{14.1}&+CLIP-KD&\textbf{26.2}&\textbf{22.2}&\textbf{29.3}&\textbf{13.7}\\ 
			\midrule
			S: EfficientNet-B0&32.6  & 27.8&40.9&20.7&S:EfficientNet-B0&32.6 & 27.8&40.9&20.7\\
			+CLIP-KD&\textbf{35.4}&\textbf{30.6}&\textbf{44.7}&\textbf{23.7}&+CLIP-KD&\textbf{34.6}&\textbf{29.4}&\textbf{44.4}&\textbf{23.1}\\ 
			\midrule
			S: ResNet-18& 28.6 & 24.0&35.3&18.1 &S:ResNet-18& 28.6 & 24.0&35.3&18.1\\
			+CLIP-KD&\textbf{31.4}&\textbf{26.9}&\textbf{39.2}&\textbf{20.0}&+CLIP-KD&\textbf{30.9}&\textbf{25.9}&\textbf{38.0}&\textbf{19.5}\\ 
			\bottomrule
	\end{tabular}}
	
	\label{vit_b_imagenet} 
	\vspace{-0.3cm}
\end{table*}

\subsection{Distilling CLIP Models}
Given the pretrained teacher CLIP model, we distill several lightweight student CLIP models with various architectures. The results are evaluated on zero-shot retrieval and ImageNet classification. \emph{We also report linear evaluation experiments on MS-COCO object detection and instance segmentation in Appendix Section 3.}

\subsubsection{Cross-Modal Retrieval on CC3M} Table~\ref{vit_b_kd_cc3m_val} reports distillation performance supervised by ViT-B/16 and ResNet-101 as teachers. The proposed KD approach improves student performance over various network architectures consistently. Supervised by ViT-B/16 for image$\to $text retrieval, KD leads to 2.0\%$\sim$3.9\% R@1 gains on MobileViT-S, Swin-T, MobileNetV3, EfficientNet-B0 and ResNet-18, respectively. For text$\to $image retrieval, KD results in 1.1\%$\sim$3.4\% R@1 gains on these networks. Supervised by ResNet-101, KD boosts the baseline by 2.0\%$\sim$4.4\% R@1 for image$\to $text retrieval, and 1.9\%$\sim$3.8\% R@1 for text$\to $image retrieval, over these five student networks, respectively. The results demonstrate the effectiveness of CLIP-KD over a series of networks. Moreover, the architectural difference between ViT and CNN does not affect CLIP-KD's performance. This is because our CLIP-KD only considers the final output embeddings for distillation instead of information from hidden layers.

\subsubsection{Zero-Shot Cross-Modal Retrieval} As shown in Table~\ref{vit_b_kd_cc3m_val}, we further transfer student CLIP models to zero-shot cross-modal retrieval on MSCOCO and Flickr.  Supervised by the teacher ViT-B/16, KD outperforms the baseline by 2.1\%$\sim$4.3\% MSCOCO R@1 margins for image$\to$text retrieval, and 1.0\%$\sim$3.3\% MSCOCO R@1 margins for text$\to$image retrieval on various networks. On Flickr, R@1 gains are 4.9\%$\sim$8.8\% for image$\to$text retrieval, and 3.2\%$\sim$6.5\% for text$\to$image retrieval.  Supervised by the teacher ResNet-101, KD leads to  1.8\%$\sim$3.7\% R@1 improvements for image$\to$text retrieval, and 1.6\%$\sim$3.6\% R@1 improvements for text$\to$image retrieval on MSCOCO. On Flickr, KD results in 3.3\%$\sim$7.8\% R@1 gains for image$\to$text retrieval, and 3.1\%$\sim$7.1\% for text$\to$image retrieval. The results reveal the transfer ability to zero-shot cross-modal retrieval using CLIP-KD.

\subsubsection{Zero-Shot ImageNet-Related Classification} In Table~\ref{vit_b_imagenet}, we transfer the student CLIP models to zero-shot ImageNet classification for visual recognition and ImageNet-variants for robustness evaluation.  For ImageNet classification supervised by ViT-B/16, KD improves 3.4\%, 3.8\%, 1.9\%, 2.8\% and 2.8\% top-1 accuracy gains over MobileViT-S, Swin-T, MobileNetV3, EfficientNet-B0 and ResNet-18, respectively. Supervised by ResNet-101, KD achieves 2.4\%, 3.1\%, 1.1\%, 2.0\% and 2.3\% top-1 accuracy improvements over five networks, respectively. The results show that CLIP-KD can help downstream visual recognition effectively. Extensive experiments over ImageNet variants indicate that CLIP-KD can lead to clear accuracy gains over baseline. 

After distillation, Swin-T even outperforms the teacher models. There are two reasons to explain this phenomenon. On the one hand, Swin-T is a powerful model, and the performance gaps with teacher models  are small. On the other hand, CLIP-KD transfers meaningful knowledge from teacher models to Swin-T, improving its performance and surpassing teacher models.

\subsubsection{Transferred from Laion-400M} 
\textbf{Cross-dataset comparison.} In Table~\ref{laion}, we use the teachers pretrained from Laion-400M~\cite{schuhmann2021laion} to distill student CLIP models trained on CC3M+12M. We find that the teacher ViT-B/16 pre-trained on Laion-400M significantly outperforms its counterpart pre-trained on CC3M+12M to distill a student ViT-T/16. It shows a 7.7\% ImageNet accuracy gain and an average cross-modal retrieval improvement of 6.8\%. The results demonstrate the CLIP-KD can effectively transfer knowledge from a large-scale dataset to improve CLIP models trained on a small-scale dataset. The advantage helps the model learn knowledge from a large-scale dataset without training too many data samples.

\textbf{Impact of teacher models with different sizes.}  In Table~\ref{laion}, we use ViT-L/14 or ViT-B/16 as two teachers to investigate the impact of teacher sizes on CLIP-KD. Both of two teachers enhance the student ViT-T/16 over baseline with substantial margins.  However, it is counter-intuitive that the more capable ViT-L/14 underperforms the weaker ViT-B/16 for distillation. One possible reason is that  a large teacher and a small student may exist capacity gaps,  making the student difficult to align with the teacher. This may become an open issue for future research. 

 \textbf{Comparison with TinyCLIP.}  CLIP-KD achieves better performance than state-of-the-art TinyCLIP~\cite{wu2023tinyclip} by 1.8\% ImageNet accuracy and 1.7\% cross-modal retrieval gains on average. Moreover, we do not provide the results of ResNet-50 for TinyCLIP, because TinyCLIP only supports the teacher and student with the same architecture-style. The results show that CLIP-KD is a more preferable method than TinyCLIP in performance and practicability.

\begin{table}[tbp]
	\centering
	\caption{\textbf{Distillation performance of zero-shot ImageNet and cross-modal retrieval trained on CC3M+12M.}  The teachers are pretrained on Laion-400M before distillation. '(from T$_{x}$)' indicates that the student is distilled from the teacher T$_{x}$.}
	\vspace{-0.2cm}
	\resizebox{1.\linewidth}{!}{
		\begin{tabular}{l|c|cc|cc}  
			\toprule
			\multirow{2}{*}{Method}&IN-1K&\multicolumn{2}{c|}{MSCOCO}&\multicolumn{2}{c}{Flickr}\\ 
			&Acc&I2T&T2I&I2T&T2I		\\	
			\midrule
			T$_{1}$: ViT-L/14 
			&72.8 &42.7&40.9&80.5 &79.5\\
			T$_{2}$: ViT-B/16 &67.1 & 39.5&36.5&76.9&75.5\\
			\midrule
			S: ViT-T/16  &30.6  & 20.7 & 20.3 & 46.4 &47.7 \\
			+TinyCLIP (from T$_{1}$)  & 39.3&26.4&24.1&57.6&57.4\\
			+TinyCLIP (from T$_{2}$)  &40.8 &26.8&24.7&58.6&58.5\\
			+CLIP-KD (from T$_{1}$)&40.9 &27.2&25.5&59.7&59.7\\
			+CLIP-KD (from T$_{2}$) &\textbf{42.6} &\textbf{28.1}&\textbf{26.0}&\textbf{60.4}&\textbf{59.9}\\
			\midrule
			S: ViT-B/16  &37.0  & 25.0 & 24.7 & 54.6 &56.6  \\
			+TinyCLIP (from T$_{1}$) & 55.4 & 35.9 & 33.6 & 73.2 & 72.8  \\
			+CLIP-KD (from T$_{1}$) &\textbf{57.5} &\textbf{37.6}&\textbf{35.6}&\textbf{75.3}&\textbf{74.5}\\
			\midrule
			S: ResNet-50 & 35.3 & 23.5 & 24.7 & 55.1 & 55.0 \\
			+CLIP-KD (from T$_{2}$) & \textbf{55.4} & \textbf{36.3} & \textbf{33.4} & \textbf{73.0} & \textbf{72.2} \\
			\bottomrule
			
	\end{tabular}}
	
	\label{laion} 
	\vspace{-0.2cm}
\end{table}

\subsection{Analysis}
In this section, we conduct thorough analyses and ablation experiments to investigate  CLIP-KD. Unless otherwise specified, the teacher and student visual encoders are
ViT-B/16 and ViT-T/16, respectively. 

\begin{figure}[t]
	\centering 
	\begin{subfigure}[t]{0.229\textwidth}
		\centering
		\includegraphics[width=\textwidth]{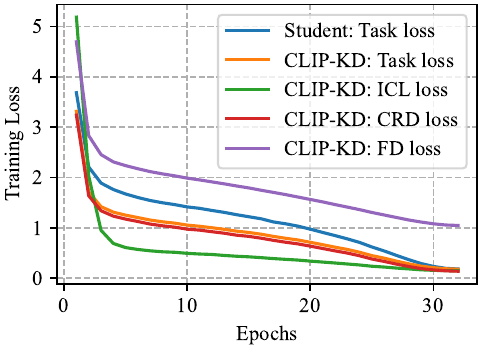}
		\caption{Training loss.}
		\label{loss_curve}
	\end{subfigure}
	\begin{subfigure}[t]{0.243\textwidth}
		\centering
		\includegraphics[width=\textwidth]{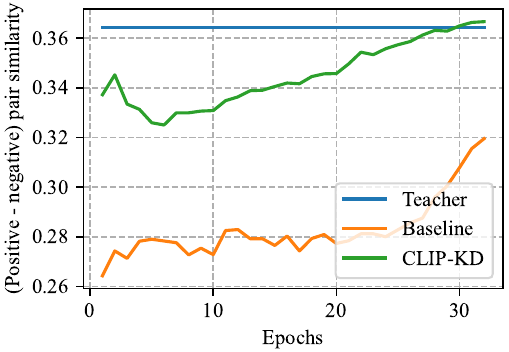}
		\caption{(Pos-neg) pair similarity.}
		\label{sim_curve}
	\end{subfigure}
	\caption{Training curves  trained on CC3M+12M  for CLIP-KD.} 
	\label{train_curvex} 
	\vspace{-0.5cm}
\end{figure}

\textbf{Training curve of CLIP-KD} As shown in Fig.~\ref{train_curvex}, we illustrate some statistics and  analyses of CLIP-KD during the training procedure: 

(1) \textbf{Training loss analysis.} Fig.~\ref{loss_curve} shows training curves of various loss terms. As the training continues, all loss values decrease and then converge until the end of training. CLIP-KD has lower task loss than that of the baseline during the training since it is supervised by a pretrained CLIP teacher. The task loss is often larger than the ICL loss, because the teacher provides converged contrastive embeddings to the student in ICL, helping the student optimize feature space readily.

(2) \textbf{Sample similarity analysis.} Fig.~\ref{sim_curve} shows the similarity curve of positive minus negative pairs, which represents the relative distance between positive and negative pairs. Contrastive learning expects positive pairs to have higher similarities while negative pairs have lower similarities. Both the baseline and CLIP-KD increase (positive-negative) pair similarity during the training stage, indicating a discriminative embedding space is gradually learned. CLIP-KD has higher similarity values than the baseline, manifesting that it guides the student to learn more discriminative features, further benefiting downstream tasks.

\begin{figure}[t]
	\centering 
	\begin{subfigure}[t]{0.49\textwidth}
		\centering
		\includegraphics[width=\textwidth]{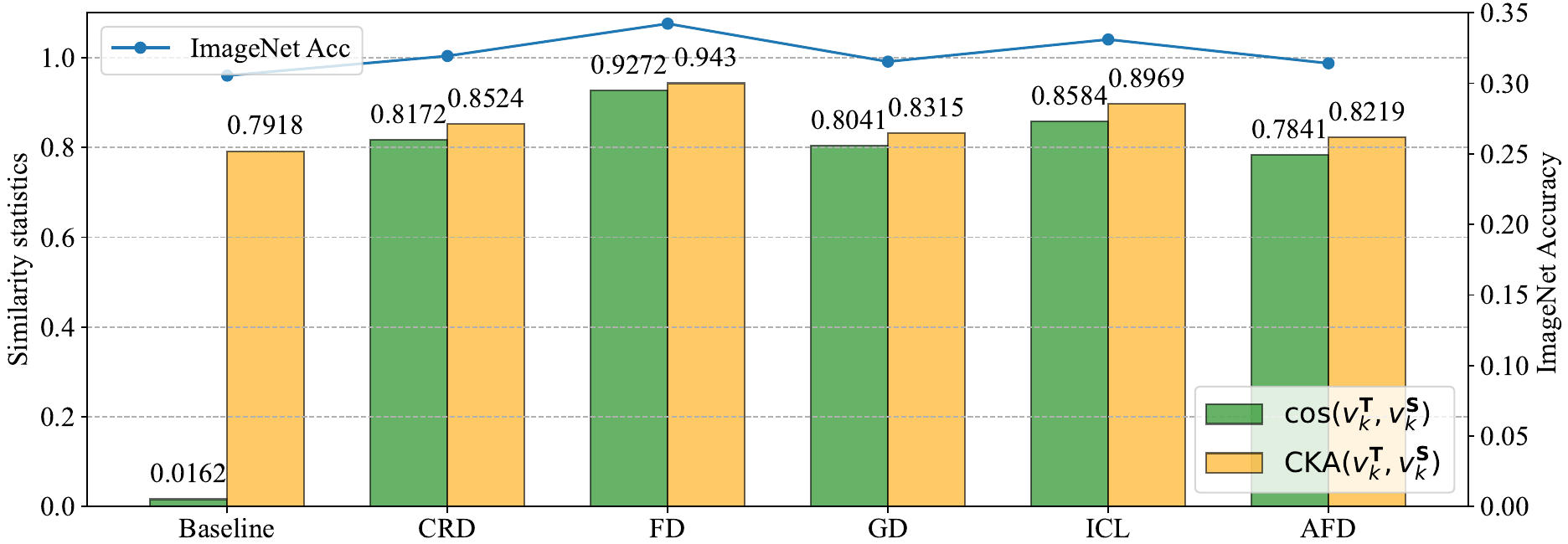}
		\caption{Similarity statistics of image features.}
		\label{img_sim}
	\end{subfigure}
	\begin{subfigure}[t]{0.49\textwidth}
		\centering
		\includegraphics[width=\textwidth]{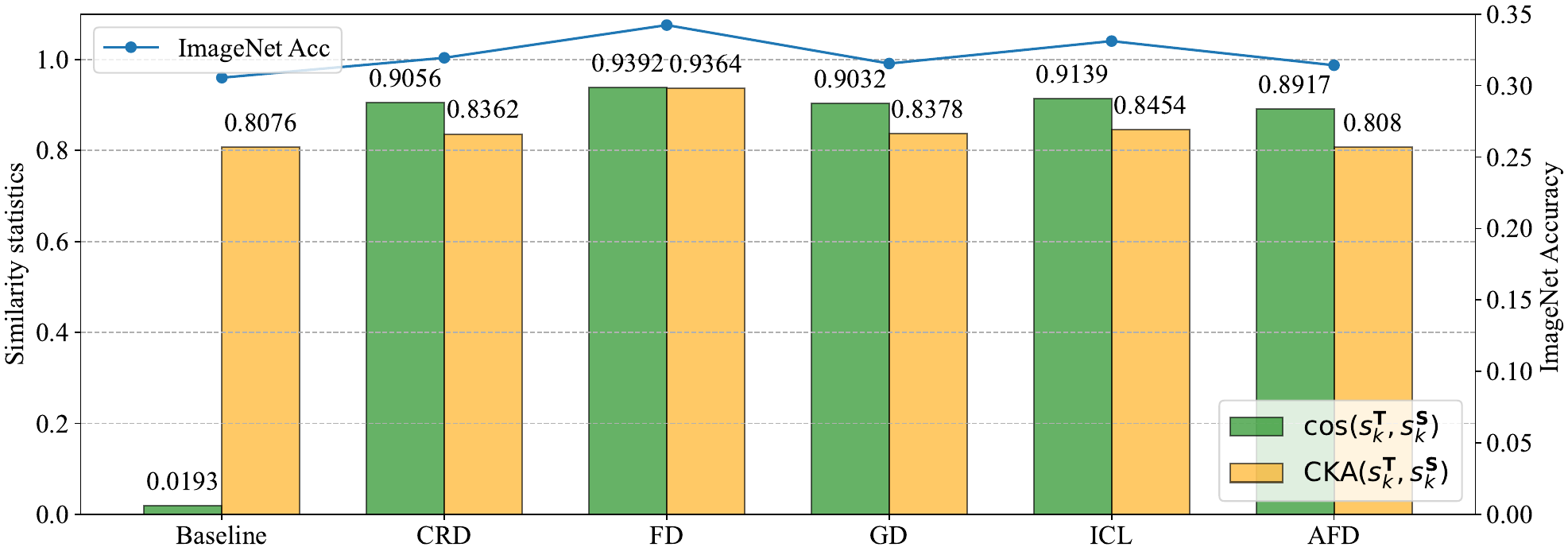}
		\caption{Similarity statistics of text features.}
		\label{txt_sim}
	\end{subfigure}
		\vspace{-0.2cm}
	\caption{\textbf{Similarity statistics between teacher and student features after distillation  trained on CC3M+12M.}  $v_{k}^{\mathbf{T}}$ and $v_{k}^{\mathbf{S}}$ denote the teacher and student image features, respectively. $s_{k}^{\mathbf{T}}$ and $s_{k}^{\mathbf{S}}$ denote the teacher and student text features, respectively. } 
	\label{img_txt_curve} 
	\vspace{-0.4cm}
\end{figure}

\textbf{Interpreting why various KD methods have different performance. } As shown in Figure~\ref{img_txt_curve}, we analyze various KD methods in different performance from the view of cosine and CKA~\cite{kornblith2019similarity} similarities between student and teacher features after distillation. We find student accuracy is in line with feature similarity. The larger similarity means that the student learns more similar teacher features, reducing the performance gap with the teacher. The simple FD performs the best because it forces the student to increase the similarity with teacher features directly. 

However, FD does not consider informative
contrastive image-text relations. ICL is proposed to promote contrastive distillation and increase mutual information between teacher and student, resulting in high similarity. By contrast, CRD, GD, and AFD are relatively weaker in enhancing similarity with the teacher, thus achieving limited gains above baseline. Overall, FD+ICL is capable of feature alignment and contrastive distillation, which is the major source of performance improvement.

\section{Conclusion}
This paper provides a comprehensive study on CLIP-KD by examining several distillation strategies, including relation, feature, gradient, and contrastive paradigms. Experimental results show that the proposed distillation methods lead to significant improvements on small CLIP models. We hope our study can provide solid CLIP-KD guidelines on practical application and attract more attention to future CLIP compression research.

\section*{Acknowledgement}
This work is partially supported by Chinese Academy of Sciences Specific Research Assistant Funding Project and Beijing Natural Science Foundation under grant 4244098. We thank Zheng Zhang from Microsoft Research Asia for helpful discussion.

{
    \small
    \bibliographystyle{ieeenat_fullname}
    \bibliography{main}
}

\appendix
\section{Gradient Distillation.}
The gradient information often shows
how the model responds to changes according to inputs. We propose to force the gradient consistency between teacher and student using the derivative w.r.t the visual and text embeddings. By this means, the student could better understand how the output should change according to the input. This helps the student behave more similarly to the teacher.

Given the image-to-text contrastive loss $\mathcal{L}_{I\to T}$, the visual embedding $v_k$ is the anchor, and the text embeddings $\{s_{b}\}_{b=1}^{|\mathcal{B}|}$ are contrastive samples. The gradient w.r.t visual and text embeddings are calculated as $\frac{\partial \mathcal{L}_{I\to T}}{\partial v_k}$ and $\frac{\partial \mathcal{L}_{I\to T}}{\partial s_b}$:
\begin{equation}
	\frac{\partial \mathcal{L}_{I\to T}}{\partial v_k}=\sum_{b=1}^{|\mathcal{B}|}(p_k[b]-\bm{1}_{[k=b]})s_b/\tau,
\end{equation}
\begin{equation}
	\frac{\partial \mathcal{L}_{I\to T}}{\partial s_b}=(p_k[b]-\bm{1}_{[k=b]})v_k/\tau.
\end{equation}
Here, $p_k$ is the contrastive distribution from $v_k$ to $\{s_{b}\}_{b=1}^{|\mathcal{B}|}$. $\bm{1}$ is an indicator function that equals to 1 when $k=b$ else returns 0. Similarly, the gradient of text-to-image contrastive loss $\mathcal{L}_{T\to I}$ w.r.t the text embedding $s_k$ and visual embeddings $\{v_{b}\}_{b=1}^{|\mathcal{B}|}$ are calculated as $\frac{\partial \mathcal{L}_{I\to T}}{\partial s_k}$ and $\frac{\partial \mathcal{L}_{I\to T}}{\partial v_b}$:
\begin{equation}
	\frac{\partial \mathcal{L}_{T\to I}}{\partial s_k}=\sum_{b=1}^{|\mathcal{B}|}(q_k[b]-\bm{1}_{[k=b]})v_b/\tau,
\end{equation}
\begin{equation}
	\frac{\partial \mathcal{L}_{T\to I}}{\partial v_b}=(q_k[b]-\bm{1}_{[k=b]})s_k/\tau.
\end{equation}
As a result, the gradient of CLIP contrastive loss $\mathcal{L}_{CLIP}$ w.r.t each visual embedding $v_k$ and text embedding $s_k$ are formulated as:
\begin{equation}
	\frac{\partial \mathcal{L}_{CLIP}}{\partial v_k}=\frac{1}{2}(\frac{\partial \mathcal{L}_{I\to T}}{\partial v_k}+\frac{\partial \mathcal{L}_{T\to I}}{\partial v_k}),
\end{equation}
\begin{equation}
	\frac{\partial \mathcal{L}_{CLIP}}{\partial s_k}=\frac{1}{2}(\frac{\partial \mathcal{L}_{I\to T}}{\partial s_k}+\frac{\partial \mathcal{L}_{T\to I}}{\partial s_k}).
\end{equation}
We align the gradient information w.r.t each visual and text embedding between teacher and student via MSE loss:
\begin{align}
	\mathcal{L}_{GD}=\frac{1}{|\mathcal{B}|}\sum_{k=1}^{|\mathcal{B}|}&(\left \| \frac{\partial \mathcal{L}_{CLIP}^{\mathbf{T}}}{\partial v_k^{\mathbf{T}}}-\frac{\partial \mathcal{L}_{CLIP}^{\mathbf{S}}}{\partial v_k^{\mathbf{S}}} \right \|  ^{2}_{2} \notag\\
	&+\left \| \frac{\partial \mathcal{L}_{CLIP}^{\mathbf{T}}}{\partial s_k^{\mathbf{T}}}-\frac{\partial \mathcal{L}_{CLIP}^{\mathbf{S}}}{\partial s_k^{\mathbf{S}}} \right \|  ^{2}_{2}).
\end{align}

\section{Theoretical Insights of Interactive Contrastive Learning: Proof of Maximizing the Lower bound of the Mutual Information}
Given the student image embedding $v_k^{\mathbf{S}}$ as the anchor and teacher text embeddings $\{s_{b}^{\mathbf{T}}\}_{b=1}^{B}$ as contrastive ones, where $B=|\mathcal{B}|$ is the batch size, the $(v_k^{\mathbf{S}}, s_{k}^{\mathbf{T}})$ is a positive pair and $\{(v_k^{\mathbf{S}}, s_{b}^{\mathbf{T}})\}_{b=1,b\ne k}^{B}$ are negative pairs. We introduce the joint distribution $\mu(v^{\mathbf{S}},s^{\mathbf{T}})$ and the product of marginals $\mu(v^{\mathbf{S}})\mu(s^{\mathbf{T}})$ . We utilize a distribution $\eta$ with an indicator variable $C$ to represent whether a pair $(v^{\mathbf{S}},s^{\mathbf{T}})$ is drawn from the joint distribution ($C=1$) or product of marginals ($C=0$):
\begin{align}
	\eta(v^{\mathbf{S}},s^{\mathbf{T}}|C=1)=\mu(v^{\mathbf{S}},s^{\mathbf{T}}), \\ \eta(v^{\mathbf{S}},s^{\mathbf{T}}|C=0)=\mu(v^{\mathbf{S}})\mu(s^{\mathbf{T}}).
\end{align} 

Here, $C=1$ represents the positive pair $(v_k^{\mathbf{S}}, s_{k}^{\mathbf{T}})$ while $C=0$ represents a negative pair from $\{(v_k^{\mathbf{S}}, s_{b}^{\mathbf{T}})\}_{b=1,b\ne k}^{B}$ , \emph{i.e.} $(v_k^{\mathbf{S}}, s_{k}^{\mathbf{T}})\sim \mu(v^{\mathbf{S}},s^{\mathbf{T}})$, $\{(v_k^{\mathbf{S}}, s_{b}^{\mathbf{T}})\}_{b=1,b\ne k}^{B}\sim \mu(v^{\mathbf{S}})\mu(s^{\mathbf{T}})$. For interactive contrastive learning, we often have $1$ positive pair for every $N$ negative pairs, where $N=B-1$. Therefore, the prior probabilities of the latent variable $C$ are formulated as:
\begin{equation}
	\eta(C=1)=\frac{1}{1+N},\  \eta(C=0)=\frac{N}{1+N}.
\end{equation}  
By using Bayes’ theorem, we can compute the class posterior of the pair $(v^{\mathbf{S}},s^{\mathbf{T}})$ belonging to the positive case ($C=1$) as :
\begin{align}
	&\eta(C=1|v^{\mathbf{S}},s^{\mathbf{T}})\\
	&=\frac{\eta(v^{\mathbf{S}},s^{\mathbf{T}}|C=1)\eta(C=1)}{\eta(v^{\mathbf{S}},s^{\mathbf{T}}|C=1)\eta(C=1)+\eta(v^{\mathbf{S}},s^{\mathbf{T}}|C=0)\eta(C=0)} \\
	&=\frac{\mu(v^{\mathbf{S}},s^{\mathbf{T}})}{\mu(v^{\mathbf{S}},s^{\mathbf{T}})+N\mu(v^{\mathbf{S}})\mu(s^{\mathbf{T}})}.
\end{align}
The log class posterior can be further expressed as follows:
\begin{align}
	& \log \eta(C=1|v^{\mathbf{S}},s^{\mathbf{T}})\\
	&=\log \frac{\mu(v^{\mathbf{S}},s^{\mathbf{T}})}{\mu(v^{\mathbf{S}},s^{\mathbf{T}})+N\mu(v^{\mathbf{S}})\mu(s^{\mathbf{T}})} \\
	&=-\log(1+N\frac{\mu(v^{\mathbf{S}})\mu(s^{\mathbf{T}})}{\mu(v^{\mathbf{S}},s^{\mathbf{T}})}) \\
	& \leq -\log(N)+\log\frac{\mu(v^{\mathbf{S}},s^{\mathbf{T}})}{\mu(v^{\mathbf{S}})\mu(s^{\mathbf{T}})}.
\end{align}

The expectations of log class posterior $\log \eta(C=1|v^{\mathbf{S}},s^{\mathbf{T}})$ can be connected to mutual information:
\begin{align}
	& \mathbb{E}_{\eta(v^{\mathbf{S}},s^{\mathbf{T}}|C=1)}\log \eta(C=1|v^{\mathbf{S}},s^{\mathbf{T}})\\
	& \leq -\log(N)+\mathbb{E}_{\mu(v^{\mathbf{S}},s^{\mathbf{T}})}\log\frac{\mu(v^{\mathbf{S}},s^{\mathbf{T}})}{\mu(v^{\mathbf{S}})\mu(s^{\mathbf{T}})}\\
	& = -\log(N)+I(v^{\mathbf{S}},s^{\mathbf{T}}),
\end{align}
where $I(v^{\mathbf{S}},s^{\mathbf{T}})$ denotes mutual information between  $v^{\mathbf{S}}$ and $s^{\mathbf{T}}$. Essentially, the ICL loss $\mathcal{L}_{ICL\_I\to T}$ is negative log class posterior of the positive pair:
\begin{equation}
	\mathcal{L}_{ICL\_I\to T}=-\log \eta(C=1|v^{\mathbf{S}},s^{\mathbf{T}}).
\end{equation}
Therefore, we can connect $	\mathcal{L}_{ICL\_I\to T}$ to the mutual information $I(v^{\mathbf{S}},s^{\mathbf{T}})$ as follows:
\begin{align}
	& \mathbb{E}_{\eta(v^{\mathbf{S}},s^{\mathbf{T}}|C=1)}\mathcal{L}_{ICL\_I\to T} \geq \log(N)-I(v^{\mathbf{S}},s^{\mathbf{T}}) \\
	\Leftrightarrow\  &I(v^{\mathbf{S}},s^{\mathbf{T}})\geq \log(N)-\mathbb{E}_{\eta(v^{\mathbf{S}},s^{\mathbf{T}}|C=1)}\mathcal{L}_{ICL\_I\to T}.
	\label{mutual_info}
\end{align}
By minimizing $	\mathcal{L}_{ICL\_I\to T}$, the lower bound on mutual information $I(v^{\mathbf{S}},s^{\mathbf{T}})$ is maximized.  The mutual information $I(v^{\mathbf{S}},s^{\mathbf{T}})$ measures uncertainty reduction in contrastive feature embeddings from the teacher text encoder when the anchor embedding from the student visual encoder is known. Since $\mathcal{L}_{ICL\_T\to I}$ is symmetric to $\mathcal{L}_{ICL\_I\to T}$, the lower bound on mutual information $I(s^{\mathbf{S}},v^{\mathbf{T}})$ can be maximized by minimizing $	\mathcal{L}_{ICL\_T\to I}$. The mutual information $I(s^{\mathbf{S}},v^{\mathbf{T}})$  measures uncertainty reduction in contrastive feature embeddings from the teacher visual encoder when the anchor embedding from the student text encoder is known.  By maximizing the lower bound of mutual information, the student network reduces uncertainty with the teacher. This means that ICL guides the student to learn more common knowledge from the teacher, leading to better feature representations.

\begin{figure}[t]
	\centering 
	\begin{subfigure}[t]{0.23\textwidth}
		\centering
		\includegraphics[width=\textwidth]{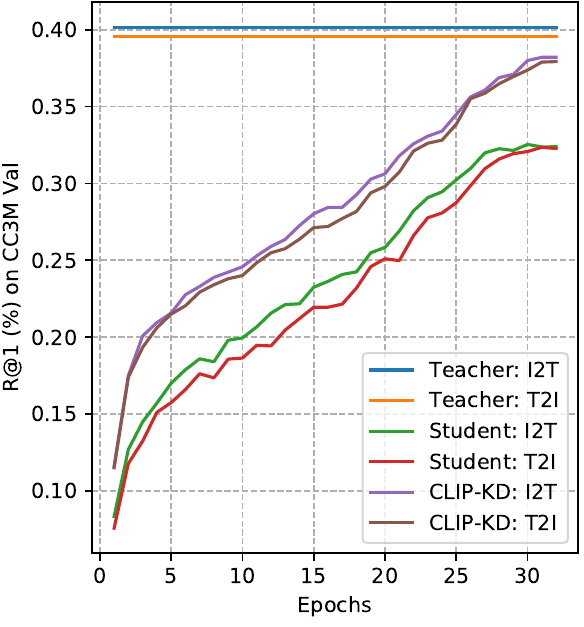}
		\caption{R@1 (\%) on CC3M Val.}
		\label{recall_curve}
	\end{subfigure}
	\begin{subfigure}[t]{0.23\textwidth}
		\centering
		\includegraphics[width=\textwidth]{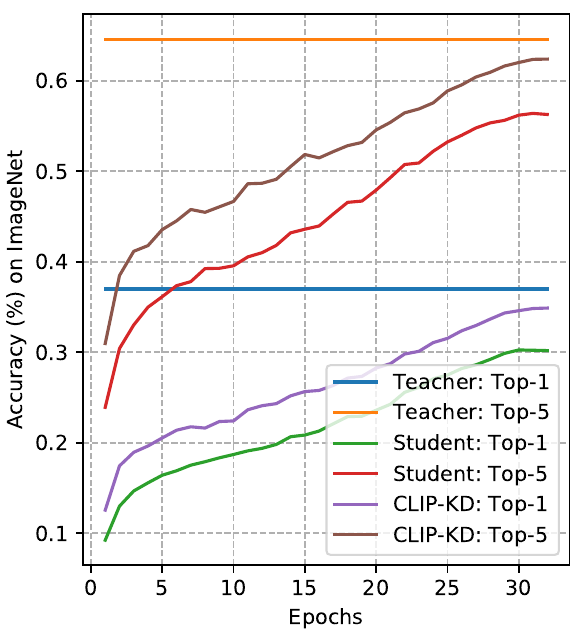}
		\caption{Accuracy (\%) on ImageNet.}
		\label{acc_curve}
	\end{subfigure}
	\caption{Training curves using ViT-B/16 as the teacher and ViT-T/16 as the student for CLIP-KD compared to the baseline.} 
	\label{train_curve} 
\end{figure}

\section{Experiments}
In this section, we conduct thorough analyses and ablation experiments to investigate  CLIP-KD. Unless otherwise specified, the teacher and student visual encoders are
ViT-B/16 and ViT-T/16, respectively.

\textbf{Analysis of Training performance curves of CLIP-KD} 
Fig.~\ref{recall_curve} and Fig.~\ref{acc_curve} show performance curves of cross-modal retrieval and ImageNet classification, respectively. CLIP-KD outperforms the baseline consistently during the training process.

\textbf{Analyses of hyper-parameters}
In this section, we investigate the impact of hyper-parameters on distillation performance.

\begin{table}[tbp]
	\centering
	\caption{Analysis of FD loss weight $\lambda_{FD}$. 'scratch$\to $converge' denotes the change of loss value from scratch to convergence. }
	
	\begin{tabular}{l|c|c|cc}  
		\toprule
		\multirow{2}{*}{$\lambda_{FD}$}&Loss&ImageNet& \multicolumn{2}{c}{CC3M Val}\\ 
		&scratch$\to $converge&Acc&I2T&T2I \\
		\midrule
		10 & 0.079$\to $0.013 & 31.1 & 33.6&33.5 \\
		100 & 0.794$\to $0.089 & 32.3  & 34.6&34.4 \\
		1000 & 7.721$\to $0.538 & 33.7 & 36.7&36.4 \\
		2000 & 15.880$\to $0.902 & \textbf{34.2} & \textbf{37.1}&\textbf{36.9} \\
		3000 & 30.452$\to $1.651 & 34.1 & 37.0&36.6 \\
		\bottomrule
	\end{tabular}
	\vspace{-0.2cm}
	\label{FD_lambda} 
\end{table}

\begin{table}[tbp]
	\centering
	\caption{Analysis of CRD loss weight $\lambda_{CRD}$. }
	
	\begin{tabular}{l|c|cc}  
		\toprule
		\multirow{2}{*}{$\lambda_{CRD}$}&ImageNet& \multicolumn{2}{c}{CC3M Val}\\ 
		&Acc&I2T&T2I \\
		\midrule
		0.5  & 31.6 &  34.9 & 34.6 \\
		1  & \textbf{31.9} &  \textbf{35.3} & \textbf{34.9} \\
		2 & 31.7 & 35.2&34.8 \\
		10& 31.2 & 34.9&34.6 \\
		\bottomrule
	\end{tabular}
	
	\label{CRD_lambda} 
\end{table}

\begin{table}[tbp]
	\centering
	\caption{Analysis of GD loss weight $\lambda_{GD}$.}
	
	\begin{tabular}{l|c|cc}  
		\toprule
		\multirow{2}{*}{$\lambda_{GD}$}&ImageNet& \multicolumn{2}{c}{CC3M Val}\\ 
		&Acc&I2T&T2I \\
		\midrule
		$10^6$ & 30.6 &  33.7 & 33.1 \\
		$10^7$  & 30.8 &  33.9 & 33.3 \\
		$10^8$ &  \textbf{31.5} & \textbf{34.5} & \textbf{34.0} \\
		$10^9$& 31.4 & 34.2 & 33.7 \\
		\bottomrule
	\end{tabular}
	
	\label{GD_lambda} 
\end{table}

\begin{table}[tbp]
	\centering
	\caption{Analysis of ICL loss weight $\lambda_{ICL}$. }
	
	\begin{tabular}{l|c|cc}  
		\toprule
		\multirow{2}{*}{$\lambda_{ICL}$}&ImageNet& \multicolumn{2}{c}{CC3M Val}\\ 
		&Acc&I2T&T2I \\
		\midrule
		0.5  & 33.7 &  37.0 & 36.8 \\
		1  & \textbf{34.2} &  \textbf{37.1} & \textbf{36.9} \\
		2 & 33.9 & 36.8&36.8 \\
		10& 33.6 & 36.3&36.3 \\
		\bottomrule
	\end{tabular}
	
	\label{ICL_lambda} 
\end{table}

\begin{figure*}[tbp]  
	\centering 
	\includegraphics[width=0.8\linewidth]{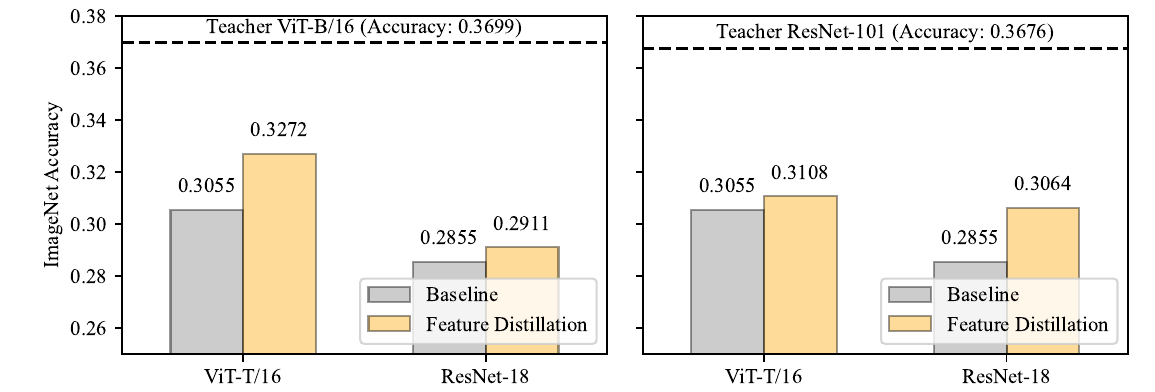}
	\caption{Top-1 accuracy on zero-shot ImageNet using intermediate feature distillation trained from CC3M+12M.}  
	\label{ifd}
	\vspace{-0.5cm}
\end{figure*}

\begin{table}[tbp]
	\centering
	\caption{Analysis of mask ratio for MFD. }
	\begin{tabular}{l|c|cc}  
		\toprule
		\multirow{2}{*}{Mask ratio}&ImageNet& \multicolumn{2}{c}{CC3M Val}\\ 
		&Acc&I2T&T2I \\
		\midrule
		0 &  \textbf{34.2} & 37.1&\textbf{36.9} \\
		0.25 &  34.1 & \textbf{37.4}&36.8  \\
		0.5 &  33.8 & 37.3 &36.7 \\
		0.75 &  33.8 & 37.1 &\textbf{36.9} \\
		\bottomrule
	\end{tabular}
	
	\label{mask_ratio} 
\end{table}

\textbf{Loss weight of FD} As shown in Table~\ref{FD_lambda}, we examine the impact of FD's loss weight $\lambda_{FD}$. The performance is gradually improved as $\lambda_{FD}$ increases but saturates at $\lambda_{FD}=2000$.

\textbf{Loss weight of CRD} As shown in Table~\ref{CRD_lambda}, we examine the impact of CRD's loss weight $\lambda_{CRD}$. Overall, the performance is robust to the weight change, where $\lambda_{CRD}=1$ is a suitable choice. This is because CRD loss is entropy-based KL-divergence loss, and the magnitude is consistent with cross-entropy-based task loss. 

\textbf{Loss weight of GD} As shown in Table~\ref{GD_lambda}, we examine the impact of GD's loss weight $\lambda_{GD}$. The performance is gradually improved as $\lambda_{GD}$ increases but saturates at $\lambda_{GD}=10^8$.

\textbf{Loss weight of ICL} As shown in Table~\ref{ICL_lambda}, we examine the impact of ICL's loss weight $\lambda_{ICL}$. Overall, the performance is robust to the weight change, where $\lambda_{ICL}=1$ achieves the best performance. ICL has the same contrastive loss function as CLIP task loss, so $\lambda_{ICL}=1$ leads to the same magnitude as CLIP task loss.

\textbf{Mask ratio} As shown in Table~\ref{mask_ratio}, we examine the impact of mask ratio. Using various mask ratios does not result in more performance gains than the no-masking baseline.

\begin{table}[t]
	\centering
	\caption{Linear evaluation on MS-COCO object detection using a CC3M+12M pretrained ResNet-50 over Mask-RCNN framework. }
	\resizebox{1.\linewidth}{!}{
		\begin{tabular}{l|cccccc}  
			\toprule
			\multirow{2}{*}{Method}&  \multicolumn{6}{c}{Object detection}  \\ 
			&AP$^{bb}$&AP$^{bb}_{50}$&AP$^{bb}_{75}$&AP$^{bb}_{S}$&AP$^{bb}_{M}$&AP$^{bb}_{L}$ \\
			\midrule
			Baseline  & 32.6 &  52.3 & 34.8 & 18.0 & 35.6 & 42.4  \\
			+CLIP-KD& \textbf{34.0} &  \textbf{53.9} & \textbf{36.5} & \textbf{20.0} & \textbf{36.8} & \textbf{43.8} \\
			\bottomrule
	\end{tabular}}
	
	\label{coco_det} 
\end{table}

\begin{table}[t]
	\centering
	\caption{Linear evaluation on MS-COCO instance segmentation using a CC3M+12M pretrained ResNet-50 over Mask-RCNN framework. }
	\resizebox{1.\linewidth}{!}{
		\begin{tabular}{l|cccccc}  
			\toprule
			\multirow{2}{*}{Method}&  \multicolumn{6}{c}{Instance segmentation}  \\ 
			&AP$^{seg}$&AP$^{seg}_{50}$&AP$^{seg}_{75}$&AP$^{seg}_{S}$&AP$^{seg}_{M}$&AP$^{seg}_{L}$ \\
			\midrule
			Baseline  &  29.9 &  49.5& 31.8& 13.1 &32.2  &44.2   \\
			+CLIP-KD& \textbf{31.1} &  \textbf{50.9} & \textbf{32.9} & \textbf{14.2} & \textbf{33.3} & \textbf{45.4} \\
			\bottomrule
	\end{tabular}}
	
	\label{coco_seg} 
\end{table}

\textbf{Distilling intermediate features. } In Figure~\ref{ifd}, we apply intermediate feature distillation across ViT and ResNet pairs. We find homogeneous pairs achieve better accuracy than heterogeneous pairs, \emph{e.g.}, ViT-T/16 obtains a 2.17\% gain supervised by ViT-B/16 but only gets a 0.56\% gain by ResNet-101. This is because the former has a more similar feature extraction process and provides student-friendly knowledge. Distilling intermediate features may be sensitive to teacher-student architectures. Therefore, we conduct the final-output-based CLIP-KD methods that use contrastive embeddings to construct distillation losses to avoid the architecture-mismatching problem.

\textbf{Linear evaluation on MS-COCO object detection and instance segmentation. } As shown in Table~\ref{coco_det}, we conduct downstream MS-COCO~\cite{lin2014microsoft} object detection and instance segmentation experiments under the same linear evaluation protocol as F-VLM~\cite{kuo2022f}. The backbone is a ResNet-50 pretrained on CC3M+12M. We adopt Mask-RCNN~\cite{he2017mask} framework, and apply the 1x training schedule to finetune the model. The implementation is based on MMDetection~\cite{chen2019mmdetection}.  We leverage the standard COCO metric Average Precision (AP) to measure performance, including bounding box detection AP (AP$^{bb}$) for object detection and mask AP (AP$^{seg}$) for instance segmentation. CLIP-KD achieves consistent performance improvements over the original CLIP without KD by average AP margins of 1.5\% and 1.2\% on object detection and instance segmentation, respectively. The results indicate that CLIP-KD can also generate better distilled features under linear evaluation for downstream dense prediction tasks.


\end{document}